\documentclass[journal]{IEEEtran}

\IEEEoverridecommandlockouts

\usepackage[noadjust]{cite}
\usepackage{amsmath,amssymb,amsfonts}
\usepackage{algorithm}
\usepackage{algpseudocode}
\usepackage{multirow}
\usepackage{graphicx}
\usepackage{textcomp}
\usepackage{xcolor}
\usepackage{caption}
\usepackage{subcaption}
\usepackage{soul}
\usepackage{comment}
\usepackage{float}
\usepackage{booktabs}

\newcommand{\reviewer}[1]{\textcolor{black}{#1}}

\begin{document}

% comment out for ArxiV version
% \begin{comment}
\makeatletter
\def\@maketitle{%
  \noindent{\footnotesize This work has been submitted to the IEEE for possible publication. Copyright may be transferred without notice, after which this version may no longer be accessible.}
  
  \begingroup
  \centering
  \vskip0.5em%
  {\Huge\@title\par}%
  \vskip1.0em%
  \@author
  \par
  \endgroup
  \vskip1em
}
\makeatother
% \end{comment}

% This code is to reduce the list of authors by using et. al:
\bstctlcite{IEEEexample:BSTcontrol}

\title{A Multi-Task Foundation Model for Wireless Channel Representation Using Contrastive and Masked Autoencoder Learning}

\author{
    \IEEEauthorblockN{
        Berkay Guler, Giovanni Geraci, and Hamid Jafarkhani
        }
\thanks{B.~Guler and H.~Jafarkhani are with the Center for Pervasive Communications and Computing, University of California, Irvine CA, USA. They were supported  in part by the NSF Award CNS-2229467.} 
\thanks{G.~Geraci is with Nokia Standards and Universitat Pompeu Fabra, Spain. He was supported in part by grants PID2021-123999OB-I00, PID2024-156488OB-I00, CEX2021-001195-M, and CNS2023-145384.}

\thanks{Some of the results in this paper will be presented at the Annual Conference
on Neural Information Processing Systems (NeurIPS), AI4NextG Workshop \cite{BGGGHJworkshop25}.}
}

\maketitle

\begin{abstract}
Current applications of self-supervised learning to wireless channel representation often borrow paradigms developed for text and image processing, without fully addressing the unique characteristics and constraints of wireless communications. %Aiming to fill 
\reviewer{To bridge }this gap, % first propose WiMAE (Wireless Masked Autoencoder), a transformer-based encoder-decoder foundation model pretrained on a realistic open-source multi-antenna wireless channel dataset. Building upon this foundation, 
\reviewer{we introduce }% develop
ContraWiMAE\reviewer{, Wireless Contrastive Masked Autoencoder, }% which enhances WiMAE by incorporating a contrastive learning objective alongside the reconstruction task in a 
\reviewer{a transformer-based foundation model that unifies masked reconstruction and masked contrastive learning for wireless channel representation. Our key innovation is a new wireless-inspired contrastive objective that exploits the inherent characteristics of wireless environment, including noise, fading, and partial observability, as natural augmentation.} %unified multi-task framework. By warm-starting from pretrained WiMAE weights and generating positive pairs via noise injection, the contrastive component enables the model to capture both structural and discriminative features, enhancing representation quality beyond what reconstruction alone can achieve. 
Through extensive evaluation on unseen scenarios \reviewer{and conditions}, we demonstrate \reviewer{our method's} effectiveness in multiple downstream tasks, \reviewer{including cross-frequency beam selection, line-of-sight detection, and channel estimation.} ContraWiMAE \reviewer{exhibits superior} linear separability and adaptability in diverse wireless environments\reviewer{, demonstrating exceptional data efficiency and competitive performance compared with supervised baselines under challenging conditions.} Comparative evaluations against a state-of-the-art wireless channel foundation model confirm the superior performance and data efficiency of our \reviewer{approach}, highlighting \reviewer{its} potential as \reviewer{a} powerful baseline for future research in self-supervised wireless channel representation learning. 
\reviewer{To foster further work in this direction, we release the model weights and training pipeline for ContraWiMAE.}
\end{abstract}

\begin{IEEEkeywords}
Wireless Channel Modeling, Foundation Models, Transformer Architectures, Masked Autoencoder, Deep Learning.
% OFDM, Self-supervised Learning
\end{IEEEkeywords}

\section{Introduction}

Large-scale self-supervised pretraining has transformed the fields of natural language processing and computer vision. This paradigm leverages diverse datasets and proxy objectives to learn broadly transferable representations, in contrast to traditional task-specific training approaches \cite{bert, gpt3, vit}. By decoupling feature learning from downstream tasks, it enables efficient, task-\reviewer{specific} adaptation. Models following this two-stage strategy---computationally intensive pretraining followed by lightweight \reviewer{adaptation}---are commonly referred to as \emph{foundation models} \cite{foundation_models}.

\subsection{Background and Motivation}

\reviewer{Traditional task-specific approaches typically require extensive labeled datasets, substantial computational resources, and dedicated model storage for each application. These constraints become particularly limiting when task-specific data are scarce or when training computational resources are restricted. Moreover, AI/ML use cases are often developed in isolation, leading to high development costs and redundant data collection efforts where overlapping information is gathered separately, causing trained models to capture similar patterns independently.}

\reviewer{Foundation models address these challenges through task-agnostic learning that employs a single backbone model to provide shared representations, with dedicated heads designed for specific use cases. This framework enables efficient adaptation to diverse downstream applications with minimal additional data and computation, as existing knowledge captured in the shared backbone can be reused across tasks \cite{FonJar2024}. The approach simplifies monitoring and management since maintaining one unified model is considerably easier than handling multiple specialized ones, and retraining or fine tuning can be accomplished within a single pipeline.}

\reviewer{This paradigm proves especially valuable in scenarios such as new frequency band deployments where historical data are limited, edge computing environments with constrained computational resources, and dynamic spectrum access systems requiring rapid adaptation to varying channel conditions. Furthermore, when complementary tasks on the user equipment side, such as beam management or channel state information (CSI) compression, share portions of the backbone architecture, task-agnostic models facilitate the integration of two-sided training or inference routines, thereby supporting enhanced model interoperability and system-wide optimization.}

\reviewer{Despite these compelling advantages, the prevailing paradigm in the wireless domain remains task-specific, with each application requiring dedicated architectures and training. The task-specific approach has shown significant potential in applications such as} channel estimation and CSI feedback compression \cite{dl_ce, dl_csi, GulJaf25, ParSohDu2025, GuoWenJin2020, WanZhaLu2025}, end-to-end communication system design \cite{e2e, YanZhuSun2024, ParSohGho2025}, and generative channel modeling \cite{XiaRanMez2022, GiuNikGer2024, LeeParKim2024}. Similarly, classification-based tasks such as beam management \cite{HenAnd2024, MaWanTia2023, KhaGabSch2023} and link blockage prediction \cite{CicDelKod2023, WuAlrCha2022} benefit from supervised deep learning techniques. These advances are also reflected in ongoing standardization efforts within 3GPP \cite{3gpp2024ai}. \reviewer{However, while these task-specific successes validate the potential of AI/ML in wireless systems, they also highlight the opportunity for more unified approaches that can leverage shared channel structure across multiple applications.}

\reviewer{We argue that the wireless communication domain presents unique opportunities for the foundation model development, as CSI inherently contains rich structural information that can benefit numerous downstream tasks. Specifically, CSI matrices serve as the fundamental information substrate that drives most critical decisions in modern communication systems, from beamforming and precoding to resource allocation and interference management. This convergence on a single input modality makes the wireless domain particularly well suited for foundation models, as learning representations directly from CSI enables knowledge transfer across diverse tasks that traditionally require separate specialized models.}

\reviewer{To capitalize on this opportunity, we propose \mbox{ContraWiMAE} \footnote{\reviewer{We make ContraWiMAE along with our dataset publicly available at:
https://github.com/BerkIGuler/WirelessContrastiveMaskedLearning}} (Wireless Masked Contrastive Autoencoder), a multi-task learning framework that introduces a novel masked contrastive objective into a wireless masked autoencoder foundation. Our approach leverages the inherent complexity of wireless channels, including noise, fading, and partial observability, as natural augmentation, contrasting differently masked versions of the same channel to learn representations that capture both structural coherence and discriminative features. In parallel with our efforts, a} limited but growing body of work has begun to explore task-agnostic self-supervised learning for wireless channel representation, with most efforts falling into two broad methodological categories: contrastive and reconstructive learning.

\emph{Contrastive learning} typically utilizes encoder-only architectures to maximize representational similarity between augmented versions of identical data points (positive pairs) while minimizing similarity between distinct samples (negative pairs) \cite{contrastive_cir_csi, contrastive_geolocation}. 
For example, one study treats the CSI and the channel impulse response (CIR) of the same wireless channel as positive pairs, maximizing their representational similarity in the embedding space \cite{contrastive_cir_csi}. A related investigation uses geospatial information by constructing a contrastive objective in which channels from identical geolocations serve as positive samples, while those from different locations serve as negative samples \cite{contrastive_geolocation}.

In parallel, \emph{reconstructive methods} aim to predict or reconstruct masked portions of the input and can be categorized into three primary architectural patterns:
\begin{enumerate}
    \item Decoder-only autoregressive models, such as GPT \cite{gpt3}.
    \item Encoder-decoder architectures, such as masked autoencoders (MAE) \cite{mae}.
    \item Encoder-only bidirectional models, such as BERT \cite{bert}.
\end{enumerate}

Among \emph{decoder-only} models, one study fine-tunes a pretrained GPT-2 language model with domain-specific adaptations for autoregressive channel prediction \cite{llm_cp}, while another extends this architecture to multimodal inputs, incorporating CSI and environmental metadata \cite{channel_gpt}. 

In terms of the \emph{encoder-decoder} approaches, the work in \cite{spectogram_vit} employs a vision transformer-based autoencoder for masked spectrogram prediction, demonstrating its effectiveness in human activity sensing and spectrogram segmentation tasks. Complementary research explores masked autoencoder pretraining for channel prediction \cite{mae_channel_pred}, although it does not address transferability across different downstream tasks.

\emph{Encoder-only} bidirectional models have received particular attention. In \cite{rf_gpt}, a transformer encoder reconstructs masked CIR samples,  although the model is described using GPT terminology despite the lack of an autoregressive decoder. A similar observation applies to \cite{wireless_gpt}, which also adopts masking-based reconstruction with an encoder-only architecture. The Large Wireless Model (LWM) represents a notable contribution, applying a masked modeling objective where random patches of multiple-input single-output (MISO) orthogonal frequency division multiplexing (OFDM) channel matrices are masked and subsequently reconstructed using a transformer encoder \cite{lwm}.

Despite these advancements, key limitations remain in current wireless foundation models. BERT-based architectures such as LWM \cite{lwm} typically employ low masking ratios (e.g., 15\%), which reduce the complexity of the pretraining task and may result in shallow or overspecialized feature representation. Furthermore, several approaches adapt BERT-style \reviewer{or GPT-style} models by replacing the original cross-entropy loss\reviewer{, designed for prediction over a discrete set,} with regression objectives tailored to continuous-valued wireless data, without thoroughly reassessing the implications for representation learning. These design choices can limit the expressiveness and generalizability of the learned representations, ultimately limiting the performance of diverse downstream wireless tasks.

More broadly, while contrastive and reconstructive approaches have each shown promise, they face fundamental limitations when applied in isolation to wireless channel modeling. Contrastive methods, for instance, are sensitive to the choice of positive and negative samples---poor sample selection can mislead the model, especially in wireless scenarios where channels may share subtle but significant similarities. These methods also typically require large batch sizes to provide a sufficient number of negative samples, increasing computational overhead \cite{cl_batch_size, cl_neg_sample_selection}. 
In contrast, reconstructive approaches tend to emphasize the accurate recovery of low-level statistical patterns, which may come at the expense of not learning the discriminative features necessary for downstream tasks such as beam selection. This produces representations that work well for regression tasks, but may underperform in classification or decision-making scenarios that depend on fine-grained feature distinctions \cite{hybrid_learning}. Combining both approaches in a multi-task framework addresses these weaknesses by enabling simultaneous learning of both structural and discriminative channel features.

\subsection{Contribution and Summary of Results}

% \reviewer{ContraWiMAE harnesses the intrinsic characteristics of wireless environments as inherent data augmentation through a novel masked contrastive objective. Rather than relying on engineered data transformations, we contrast masked views of identical channel realizations, enabling the model to identify invariant propagation characteristics across different partial observations while preserving both the underlying channel structure and discriminative capacity essential for diverse downstream applications.}

\reviewer{Our approach} directly addresses the limitations of prior reconstructive approaches by adopting an asymmetric encoder-decoder architecture that emphasizes abstraction over memorization. Inspired by techniques in the vision domain \cite{beit, mae}, \reviewer{we utilize the} regression loss with a structure that more effectively encourages the extraction of semantically meaningful features \reviewer{compared to encoder-only and decoder-only architectures that naively substitute regression for classification losses}. Specifically \reviewer{in our approach}, \reviewer{the} encoder only processes the visible patches of the input, while a lightweight decoder reconstructs the masked portions \reviewer{from the latent representations of visible patches}. 

% Unlike LWM’s shallow masking, WiMAE employs higher masking ratios, forcing the model to learn robust representations from limited observations and resulting in more transferable features across various wireless tasks. 

% Building upon the WiMAE foundation, we further enhance our approach by developing ContraWiMAE, which incorporates a contrastive learning objective alongside the reconstruction task in a unified multi-task framework. By warm-starting from pretrained WiMAE weights and generating positive pairs via the injection of additive white Gaussian noise (AWGN), ContraWiMAE contrasts these against negative samples drawn from different channel realizations. This enables the model to learn representations that are not only effective for reconstruction but also highly discriminative. The resulting hybrid approach significantly increases the linear separability of the learned embeddings, allowing simpler downstream models to achieve better performance without relying on complex architectures to compensate for representational weaknesses.

\reviewer{This reconstruction-only foundation serves as both a strong baseline for comparison and a critical building block for our complete framework. Building upon this reconstruction-only foundation, we introduce a novel masked contrastive objective and continually pretrain with a multi-task objective. ContraWiMAE, instead of requiring engineered augmentations or heuristic-based positive sample selection, generates contrastive pairs by applying different random masks to noisy versions of identical channel realizations. The key innovation lies in teaching the model to recognize that different partial views of the same propagation environment should map to similar representations, while maintaining discriminative boundaries between fundamentally different channels. This masked contrastive objective enables the extraction of invariant features that remain consistent across varying noise conditions and measurement patterns, which are critical properties for real wireless deployments where only noisy and partial channel observations are available. Furthermore, ContraWiMAE preserves the strong reconstruction capabilities through multi-task learning while significantly enhancing the linear separability of learned embeddings.}

\reviewer{We evaluate our approach on three complementary downstream tasks that represent different categories of wireless applications: cross-frequency beam selection (a complex classification task requiring detailed spatial understanding), line-of-sight detection (a simple classification task for coarse channel characterization), and channel estimation (a reconstruction task leveraging learned channel structure). Although our current evaluation focuses on these representative tasks, the task-agnostic nature of our learned representations provides a foundation for broader applicability to additional wireless communication challenges.} 

Key findings from our experiments include the following: \reviewer{
\begin{itemize} 
\item Our dual masked contrastive and reconstructive learning objective enables competitive performance with supervised training across tasks. 
\item Masked contrastive learning using noise as augmentation improves robustness to low-SNR conditions and distribution shifts across frequencies and environments. 
\item The masked contrastive objective promotes a linearly separable and locally structured feature space, demonstrated by linear probing and nearest-neighbor evaluations. 
\item ContraWiMAE enables simple downstream models to match the performance of complex networks, enabling practical wireless adoption. 
\item Our dual learning strategy maintains reconstruction capabilities with minimal degradation compared to reconstruction-only learning.
\end{itemize}}

\subsubsection*{Paper organization}

The remainder of this paper is organized as follows. Section~\ref{sec:sys_model} describes the system model and problem formulation, together with the training objectives used for model optimization. Section~\ref{sec:architecture} presents the ContraWiMAE architecture, detailing the preprocessing steps, encoder and decoder designs, and the integration of contrastive learning. Section~\ref{sec:sim} outlines the methodology, including dataset generation, pretraining setup, and adaptations for downstream tasks. \reviewer{Section~\ref{sec:complexity} includes the analysis of the computational complexity of our method.} Section~\ref{sec:results} reports and thoroughly analyzes the experimental results, highlighting their implications for wireless representation learning. Finally, Section~\ref{sec:conclusion} concludes the paper.

\section{System Model and Problem Formulation}
\label{sec:sys_model}

In this section, we describe the system model and problem formulation, along with the training objectives used for model optimization.

\subsection{System Model} 
\reviewer{We consider a MISO-OFDM system with $N_{\text{f}}$ subcarriers and $\Delta f$ kHz spacing where the base station utilizes a uniform linear array (ULA) with $N_{\text{s}}$ antennas. The user equipment employs a single omnidirectional antenna. The channel response in the frequency domain for each symbol period is denoted by $\mathbf{H}\in \mathbb{C}^{N_{\text{s}} \times N_{\text{f}}}$ and represents the channel between the base station and the users.} \reviewer{The assumption of available channel state information follows standard practice in machine learning for wireless communications research, where CSI is typically obtained through pilot-based channel estimation in contemporary systems like LTE and 5G.} 
To perform a fair comparison, our system model parameters are purposely chosen to match the configuration used in LWM \cite{lwm}.  

\subsection{Problem Formulation}

We begin by formulating the \reviewer{reconstruction} approach, then extend it to the ContraWiMAE framework that incorporates \reviewer{our masked contrastive objective.} \reviewer{Our unconstrained pretraining objectives are designed to learn generic representations without bias toward specific constraint sets, enabling task-specific constraints to be incorporated during downstream application phases.}

\subsubsection{Masked Reconstruction Objective}

We model the problem with a parameterized encoder-decoder pair $(f_\theta, g_\phi)$. Given a complex channel matrix $\mathbf{H} \in \mathbb{C}^{N_{\text{s}} \times N_{\text{f}}}$, our aim is to learn an efficient representation that captures the underlying structure with only partial observations. \reviewer{We employ random masking rather than structured patterns to learn general-purpose representations that adapt to various loss patterns without bias toward specific failure modes. This promotes robust feature learning that transfers to actual structured losses during downstream applications.} 

Let ${\mathbf{M}^{\prime}}$ be a random masking operator drawn from a distribution $p(\mathbf{M}^{\prime})$ that removes a portion of the information from $\mathbf{H}$, resulting in a partial observation $\mathbf{H}_{\text{m}} = {\mathbf{M}^{\prime}}(\mathbf{H})$. Also, let $\overline{{\mathbf{M}}}^{\prime}$ denote the complementary mask operation that selects only the masked portions.

We jointly optimize $f_\theta$ and $g_\phi$ to reconstruct the masked portions of $\mathbf{H}$ from the visible portions $\mathbf{H}_{\text{m}}$ by minimizing the reconstruction error in the minimum squared error (MSE) sense, focusing only on the masked regions. The loss function $\mathcal{L}_{\text{recon}}$ is defined as:
\begin{equation}
\begin{split}
\mathcal{L}_{\text{recon}}(\theta, \phi) = \mathbb{E}_{\mathbf{H} \sim p(\mathbf{H}), \mathbf{M}^\prime \sim p(\mathbf{M}^\prime)} \Big[ 
\| \overline{{\mathbf{M}}}^{\prime}(\mathbf{H}) - \\
\overline{{\mathbf{M}}}^{\prime}(g_\phi(f_\theta(\mathbf{H}_{\text{m}}))) \|_F^2 \Big],
\end{split}
\label{eq:recon_loss}
\end{equation}
where $\|\cdot\|_F$ denotes the Frobenius norm and $p(\mathbf{H})$ represents the distribution of channel matrices.

\subsubsection{ContraWiMAE: Hybrid Reconstruction and Contrastive Learning}

We enhance our approach by incorporating a contrastive learning objective alongside the reconstruction task in a multi-task framework.

\subsubsection*{Positive pair generation}
For each channel matrix $\mathbf{H}_i$, we create a positive pair $\mathbf{H}_i^+$ by adding AWGN:
\begin{equation}
\mathbf{H}_i^+ \sim \mathcal{CN}(\mathbf{H}_i, \sigma^2\mathbf{I}),
\end{equation}
where $\sigma^2$ is a hyperparameter that controls the noise variance.

\subsubsection*{Contrastive learning objective}
Let $\mathbf{H}_{i,\text{m}} = \mathbf{M}\reviewer{_i}^\prime(\mathbf{H}_i)$ and $\mathbf{H}_{i,\text{m}}^+ = \mathbf{M}\reviewer{_i}^{\reviewer{+}\prime}(\mathbf{H}_i^+)$ represent the masked versions of the original and augmented channel matrices, respectively\reviewer{, where $\mathbf{M}_i^\prime$ and $\mathbf{M}_i^{+\prime}$ are different random masks}. 
Let $\mathbf{z}_i = \mathcal{G}_{\psi}(f_\theta(\mathbf{H}_{i,\text{m}}))$ and $\mathbf{z}_i^+ = \mathcal{G}_{\psi}(f_\theta(\mathbf{H}_{i,\text{m}}^+))$ be the global embeddings derived by applying a nonlinear operator $\mathcal{G}_{\psi}$ parameterized by $\psi$, which generates a low-dimensional representation of the encoder output. These embeddings are normalized to lie on a unit hypersphere for contrastive loss calculation. \reviewer{We concatenate representations from both views across the batch to form $\mathbf{Z} = [\mathbf{z}_1, \mathbf{z}_2, \ldots, \mathbf{z}_{\mathcal{B}}, \mathbf{z}_1^+, \mathbf{z}_2^+, \ldots, \mathbf{z}_{\mathcal{B}}^+] \in \mathbb{R}^{2\mathcal{B} \times d_\text{c}}$, where $d_\text{c}$ is the contrastive space dimension.} Inspired by Noise Contrastive Estimation \cite{nce} and following the InfoNCE approach \cite{infonce}, we formulate \reviewer{our masked} contrastive objective as:

\begin{equation}
\begin{split}
\mathcal{L}_{\text{contra}}(\theta, \psi) &= \reviewer{-\frac{1}{2\mathcal{B}} \sum_{k=1}^{2\mathcal{B}} \log \frac{\exp(\mathbf{z}_k \cdot \mathbf{z}_{\text{pos}(k)} / \tau)}{\sum_{j=1, j \neq k}^{2\mathcal{B}} \exp(\mathbf{z}_k \cdot \mathbf{z}_j / \tau)}},
\end{split}
\label{eqn:contra_objective}
\end{equation}
\reviewer{where $\mathbf{z}_k$ denotes the $k$-th row of $\mathbf{Z}$, $\text{pos}(k) = k + \mathcal{B}$ if $k \leq \mathcal{B}$, otherwise $\text{pos}(k) = k - \mathcal{B}$,} $\tau$ is a temperature parameter, and $\mathcal{B}$ is the batch size. % This loss encourages the representation of a channel and its augmented version to be similar, while pushing apart representations of different channels in the same batch.

\subsubsection*{Combined multi-task objective}
The combined training objective becomes:
\begin{equation}
\mathcal{L}(\theta, \phi, \psi) = \alpha \mathcal{L}_{\text{recon}}(\theta, \phi) + (1-\alpha) \mathcal{L}_{\text{contra}}(\theta, \psi),
\label{eqn:combined_objective}
\end{equation}
where $\alpha$ is a coefficient that trades off the reconstruction and contrastive objectives.
The optimal parameters $\theta^*$, $\phi^*$, and $\psi^*$ are obtained through:
\begin{equation}
(\theta^*, \phi^*, \psi^*) = \underset{\theta, \phi, \psi}{\arg\min} \; \mathcal{L}(\theta, \phi, \psi).
\end{equation}
\subsubsection{Model Utilization Strategy}

While our primary focus is on the encoder $f_{\theta}$, which produces the latent representations used for downstream tasks, both the decoder $g_{\phi}$ and the contrastive learning component $\mathcal{G}_{\psi}$ play critical roles during training. The decoder enforces the reconstruction objective, encouraging the encoder to extract structurally informative features. The \reviewer{masked} contrastive objective \reviewer{favors representations that are invariant to particular antennas and subcarriers while remaining noise invariant. This prevents overfitting to noise artifacts and specific measurement patterns, while maintaining semantic consistency across partial observations.} 

This multi-task training approach allows ContraWiMAE to simultaneously learn both structural and discriminative features, leading to more robust representations that generalize better across diverse wireless environments. During inference, the decoder \reviewer{and contrastive head are} typically discarded when only the encoder's embeddings are needed (e.g., for downstream classification tasks). \reviewer{Therefore, the decoder complexity may not pose as serious implications as encoder complexity for inference time and compute resources.} However, for tasks that require accurate signal reconstruction, the decoder remains relevant and can be retained to generate full outputs from partial observations.

\section{Proposed Architecture}
\label{sec:architecture}

\begin{figure*}[!t]
    \centering
    \includegraphics[width=\textwidth]{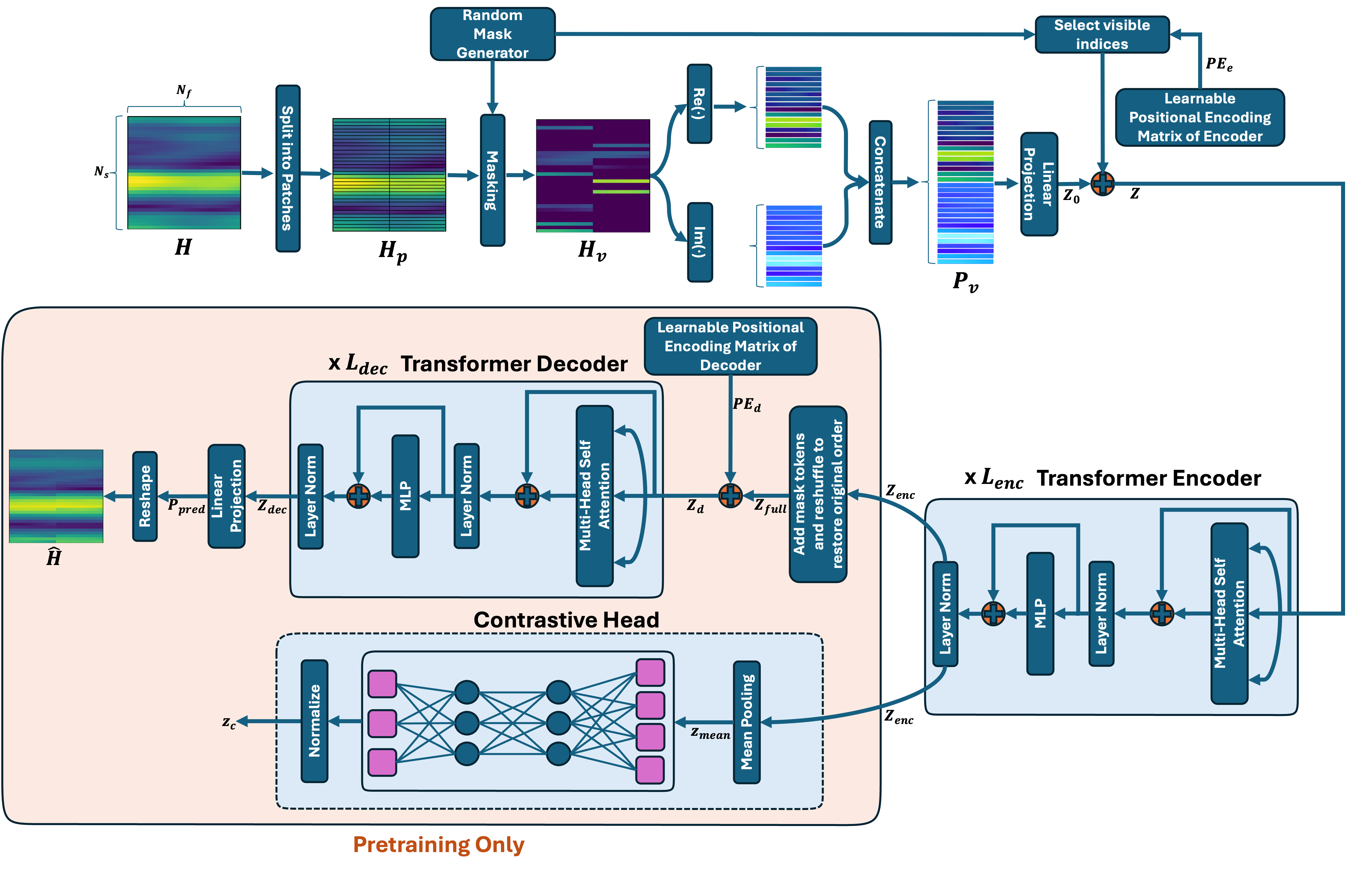}
    \caption{Illustration of the ContraWiMAE \reviewer{blocks}, detailing preprocessing steps, encoder\reviewer{-decoder} processing, and \reviewer{the contrastive head}.}
    \label{fig:arch}
\end{figure*}

In this section, we detail our proposed architecture to model the encoder-decoder pair $(f_\theta, g_\phi)$ and the contrastive operator $\mathcal{G}_{\psi}$. The main building blocks are illustrated in Fig.~\ref{fig:arch}, in which the contrastive head block, pictured with dashed lines, is the only difference between \reviewer{reconstruction-only baseline} and ContraWiMAE.

\subsection{Preprocessing Operations}

\subsubsection*{Partitioning}
We partition the channel matrix $\mathbf{H} \in \mathbb{C}^{N_{\text{s}} \times N_{\text{f}}}$ into a set of non-overlapping patches, each with dimension $d_{\text{p}}= N_{\text{p,s}} \times N_{\text{p,f}}$. 
%with $d_p=N_{p,s}N_{p,f}$. 
For simplicity, we assume that $N_{\text{s}}$ is divisible by $N_{\text{p,s}}$ and $N_{\text{f}}$ is divisible by $N_{\text{p,f}}$. This partitioning operation can be expressed as:
\begin{equation}
\begin{split}
\mathbf{H}_{\text{p}} = \{\mathbf{p}_{i,j} \in \mathbb{C}^{N_{\text{p,s}} \times N_{\text{p,f}}} \mid %\\
1 \leq i \leq \frac{N_{\text{s}}}{N_{\text{p,s}}}, 1 \leq j \leq \frac{N_{\text{f}}}{N_{\text{p,f}}}\},
\end{split}
\end{equation}
where $\mathbf{p}_{i,j}$ represents the patch at position $(i,j)$ in the partitioned space.

\subsubsection*{Masking}

We then apply a binary mask to these patches. Let $\mathbf{M} \in \{0,1\}^{\frac{N_{\text{s}}}{N_{\text{p,s}}} \times \frac{N_{\text{f}}}{N_{\text{p,f}}}}$ be a mask that operates on patches, with entries $M_{i,j}$. We only retain visible patches, i.e., those for which $M_{i,j} = 1$, as follows:
\begin{equation}
\mathbf{H}_{\text{v}} = \{\mathbf{p}_{i,j} \mid \mathbf{p}_{i,j} \in \mathbf{H}_{\text{p}}, M_{i,j} = 1\}.
\end{equation}
The mask ratio $m_{\text{r}}$, defined as the proportion of patches removed from processing, is calculated as:
\begin{equation}
m_{\text{r}} = 1 - \frac{1}{K} \sum_{i=1}^{\frac{N_{\text{s}}}{N_{\text{p,s}}}}\sum_{j=1}^{\frac{N_{\text{f}}}{N_{\text{p,f}}}} M_{i,j},
\end{equation}
where $K = \frac{N_{\text{s}}}{N_{\text{p,s}}} \cdot \frac{N_{\text{f}}}{N_{\text{p,f}}}$ is the total number of patches. 

\subsubsection*{Vectorization}
Next, we transform the visible patches into a sequence representation by ordering them based on their original positions in a row-major order. Let $\mathbf{p}_{\text{v}}^k$ be the $k$-th visible patch in this ordering, with $1 \leq k \leq N_{\text{v}}$, where $N_{\text{v}} = (1-m_{\text{r}}) \cdot K$ is the number of visible patches.
To handle the complex-valued nature of the channel matrix, we separate each patch into its real and imaginary components \reviewer{rather than magnitude and phase to preserve linearity and avoid discontinuities at phase wrapping boundaries.} We define $(\mathbf{p}_{\text{v}}^k)^{\text{r}} = \Re(\mathbf{p}_{\text{v}}^k)$ and $(\mathbf{p}_{\text{v}}^k)^{\text{i}} = \Im(\mathbf{p}_{\text{v}}^k)$, respectively, each with dimension $N_{\text{p,s}} \times N_{\text{p,f}}$. We flatten each component to a vector of length $d_{\text{p}}$ and construct the visible patch sequence:
\begin{equation}
\begin{aligned}
\mathbf{P}_{\text{v}} = &[\text{vec}((\mathbf{p}_{\text{v}}^1)^{\text{r}}), \ldots, \text{vec}((\mathbf{p}_{\text{v}}^{N_{\text{v}}})^{\text{r}}), \\
&\text{vec}((\mathbf{p}_{\text{v}}^1)^{\text{i}}), \ldots, \text{vec}((\mathbf{p}_{\text{v}}^{N_{\text{v}}})^{\text{i}})]^T \in \mathbb{R}^{2N_{\text{v}} \times d_{\text{p}}},
\end{aligned}
\label{eq:vectorization}
\end{equation}
where $\text{vec}(\cdot)$ denotes the vectorization operation that transforms a matrix into a column vector.

\subsubsection*{Linear projection}
The sequence of visible patches $\mathbf{P}_{\text{v}}$ is first processed through a linear projection layer, where a projection matrix $\mathbf{W}_0 \in \mathbb{R}^{d_{\text{p}} \times d_{\text{e}}}$ transforms each patch from its original dimension $d_{\text{p}}$ to the hidden dimension of the encoder $d_{\text{e}}$:
\begin{equation}
\mathbf{Z}_0 = \mathbf{P}_{\text{v}}\mathbf{W}_0 + \mathbf{B}_0,
\label{eq:linear_proj}
\end{equation}
where $\mathbf{B}_0 \in \mathbb{R}^{2N_{\text{v}} \times d_{\text{e}}}$ is the bias matrix.

\subsubsection*{Positional encoding}
To incorporate positional information, we maintain a set of positional encodings $\mathbf{PE}_{\text{e}} \in \mathbb{R}^{2K \times d_{\text{e}}}$ for all possible patch positions. The first $K$ rows correspond to the positions of real components while the last $K$ rows correspond to the positions of imaginary components. Let $\mathcal{I}_{\text{v}} \subset \{1, 2, ..., 2K\}$ be the set of indices in the original patch sequence corresponding to visible patches (both real and imaginary components). We add the subset of positional encodings corresponding to these visible patches as follows:
\begin{equation}
\mathbf{Z} = \mathbf{Z}_0 + \mathbf{PE}_{\text{e}}[\mathcal{I}_{\text{v}}, :],
\label{eq:pos_enc}
\end{equation}
where $\mathbf{PE}_{\text{e}}[\mathcal{I}_{\text{v}}, :]$ denotes the rows of $\mathbf{PE}_{\text{e}}$ indexed by $\mathcal{I}_{\text{v}}$. This ensures that the network's input $\mathbf{Z} \in \mathbb{R}^{2N_{\text{v}} \times d_{\text{e}}}$ retains both the content information of the visible patches and their spatial positions within the original channel matrix.

\subsection{Encoder Architecture}

The encoder consists of $L_{\text{enc}}$ transformer layers as described in what follows. Let $\mathbf{Z}^{(0)} = \mathbf{Z} \in \mathbb{R}^{2N_{\text{v}} \times d_{\text{e}}}$ be the initial input to the first layer.

\subsubsection*{Multi-head self-attention}
Each layer $l \in \{1, 2, \ldots, L_{\text{enc}}\}$ takes the output of the previous layer $\mathbf{Z}^{(l-1)}$ and processes it through multi-head self-attention (MHSA) with $M_{\text{enc}}$ attention heads. 
For any layer, given an input $\mathbf{Z}_{\text{in}}$, the scaled dot-product attention in each head $i$ is computed as $\text{head}_i = \mathbf{A}_i(\mathbf{Z}_{\text{in}}\mathbf{W}_i^{\text{V}} + \mathbf{B}_i^{\text{V}})$, where $\mathbf{A}_i$ is defined as 
\begin{equation}
\mathbf{A}_i = \text{softmax}\left(\frac{(\mathbf{Z}_{\text{in}}\mathbf{W}_i^{\text{Q}} + \mathbf{B}_i^{\text{Q}})(\mathbf{Z}_{\text{in}}\mathbf{W}_i^{\text{K}} + \mathbf{B}_i^{\text{K}})^\top}{\sqrt{d_{\text{e}}/M_{\text{enc}}}}\right),
\label{eq:attention}
\end{equation}
where $\mathbf{W}_i^{\text{Q}}$, $\mathbf{W}_i^{\text{K}}$, $\mathbf{W}_i^{\text{V}} \in \mathbb{R}^{d_{\text{e}} \times (d_{\text{e}}/M_{\text{enc}})}$ are the query, key, and value projection matrices, respectively, and $\mathbf{B}_i^{\text{Q}}$, $\mathbf{B}_i^{\text{K}}$, $\mathbf{B}_i^{\text{V}} \in \mathbb{R}^{2N_{\text{v}} \times (d_{\text{e}}/M_{\text{enc}})}$ are the corresponding bias matrices for the $i$-th head. Here, $\mathbf{A}_i \in \mathbb{R}^{2N_{\text{v}} \times 2N_{\text{v}}}$ represents the attention weights that capture the relationships between visible patches. 

% The $M_{\text{enc}}$ attention heads are concatenated and linearly projected to form the MHSA output.

\subsubsection*{Multi-head concatenation and projection}
The outputs of all $M_{\text{enc}}$ attention heads are concatenated and linearly projected to obtain the final MHSA output:
\begin{equation}
\begin{aligned}
\text{MHSA}(\mathbf{Z}_{\text{in}}) = &\text{Concat}(\text{head}_1, \text{head}_2, \\
&\ldots, \text{head}_{M_{\text{enc}}})\mathbf{W}^{\text{O}} + \mathbf{B}^{\text{O}},
\end{aligned}
\label{eq:mhsa}
\end{equation}
where $\text{Concat}(\text{head}_1, \text{head}_2, \ldots, \text{head}_{M_{\text{enc}}}) \in \mathbb{R}^{2N_{\text{v}} \times d_{\text{e}}}$ is the concatenation of all attention heads, $\mathbf{W}^{\text{O}} \in \mathbb{R}^{d_{\text{e}} \times d_{\text{e}}}$ is the output projection matrix, and $\mathbf{B}^{\text{O}} \in \mathbb{R}^{2N_{\text{v}} \times d_{\text{e}}}$ is the output bias.

\subsubsection*{Residual connection, layer normalization, and feed-forward network}
The transformer layer applies residual connections, layer normalization (LN), and a multilayer perceptron (MLP):
\begin{equation}
\mathbf{Z}_{\text{mid}} = \text{LN}(\text{MHSA}(\mathbf{Z}_{\text{in}}) + \mathbf{Z}_{\text{in}}),
\end{equation}
\begin{equation}
\mathbf{Z}_{\text{out}} = \text{LN}(\text{MLP}(\mathbf{Z}_{\text{mid}}) + \mathbf{Z}_{\text{mid}}),
\label{eq:mlp}
\end{equation}
where $\text{MLP}(\mathbf{Z}_{\text{mid}}) = \text{GELU}(\mathbf{Z}_{\text{mid}}\mathbf{W}_1 + \mathbf{B}_1)\mathbf{W}_2 + \mathbf{B}_2$ with $\mathbf{W}_1 \in \mathbb{R}^{d_{\text{e}} \times 2d_{\text{e}}}$ and $\mathbf{W}_2 \in \mathbb{R}^{2d_{\text{e}} \times d_{\text{e}}}$ as projection matrices, $\mathbf{B}_1 \in \mathbb{R}^{2N_{\text{v}} \times 2d_{\text{e}}}$ and $\mathbf{B}_2 \in \mathbb{R}^{2N_{\text{v}} \times d_{\text{e}}}$ as bias matrices, and GELU as the Gaussian Error Linear Unit activation function \cite{gelu}.

For each layer $l$, the output $\mathbf{Z}^{(l)} = \mathbf{Z}_{\text{out}}$ becomes the input to the next layer: $\mathbf{Z}_{\text{in}} = \mathbf{Z}^{(l)}$ for layer $l+1$. 
After passing through all transformer layers, the final output $\mathbf{Z}^{(L_{\text{enc}})}$ becomes the encoder output $\mathbf{Z}_{\text{enc}} \in \mathbb{R}^{2N_{\text{v}} \times d_{\text{e}}}$ that captures contextual information from visible patches while maintaining their positional relationships.

\subsection{Decoder Architecture}

The decoder reconstructs the full channel matrix by processing both visible and masked patches. During pretraining, we first recover the complete patch sequence by restoring the original ordering of visible patches and inserting learnable mask tokens for the masked patches.

\subsubsection*{Restoring patch ordering and inserting learnable mask token}
We initialize a sequence $\mathbf{Z}_{\text{full}} \in \mathbb{R}^{2K \times d_{\text{e}}}$ to represent all patches (both visible and masked) in their original positions. For each position $i$ in the full sequence:
\begin{equation}
\mathbf{Z}_{\text{full}}[i] = 
\begin{cases}
\mathbf{Z}_{\text{enc}}[j], & \text{if } i \in \mathcal{I}_{\text{v}} \text{ and } \mathcal{I}_{\text{v}}[j] = i \\
\mathbf{m}_{\text{mask}}, & \text{if } i \notin \mathcal{I}_{\text{v}}
\end{cases}
\label{eq:z_full}
\end{equation}
where $\mathcal{I}_{\text{v}}[j]$ denotes the original position index of the $j$-th visible patch and $\mathbf{m}_{\text{mask}} \in \mathbb{R}^{d_{\text{e}}}$ is a shared, learnable mask token that is initialized as a zero vector but updated during training via backpropagation. This allows the model to learn an optimal representation for the masked tokens rather than being constrained to zeros.

\subsubsection*{Positional encoding}
To incorporate positional information in the decoder, we add positional encodings $\mathbf{PE}_{\text{d}} \in \mathbb{R}^{2K \times d_{\text{e}}}$ directly to the sequence:
\begin{equation}
\mathbf{Z}_{\text{d}} = \mathbf{Z}_{\text{full}} + \mathbf{PE}_{\text{d}},
\end{equation}
where $\mathbf{PE}_{\text{d}}$ represents the positional encodings corresponding to the original positions of all patches in the sequence.

\subsubsection*{Transformer layers}
The decoder consists of $L_{\text{dec}}$ transformer layers, each with $M_{\text{dec}}$ attention heads. Each layer employs the same architecture as the encoder's transformer layers, with multi-head self-attention, residual connections, layer normalization, and multilayer perceptrons. The decoder uses the same dimension $d_{\text{e}}$ as the encoder throughout its layers. 
The attention mechanism of the decoder allows both visible and masked patches to attend to each other, enabling information flow from visible patches to masked ones for reconstruction.

\subsubsection*{Final projection and reshaping}
After passing through all transformer layers, the decoder produces output features $\mathbf{Z}_{\text{dec}} \in \mathbb{R}^{2K \times d_{\text{e}}}$ for all patches. These are then projected to the original patch dimension through a linear layer:
\begin{equation}
\mathbf{P}_{\text{pred}} = \mathbf{Z}_{\text{dec}}\mathbf{W}_{\text{out}} + \mathbf{B}_{\text{out}},
\label{eq:final_proj}
\end{equation}
where $\mathbf{W}_{\text{out}} \in \mathbb{R}^{d_{\text{e}} \times d_{\text{p}}}$ and $\mathbf{B}_{\text{out}} \in \mathbb{R}^{2K \times d_{\text{p}}}$ are the output projection matrix and bias matrix, respectively.
The first $K$ rows of $\mathbf{P}_{\text{pred}}$ correspond to the real components, while the last $K$ rows correspond to the imaginary components of the reconstructed patches. 
The predicted patches $\mathbf{P}_{\text{pred}} \in \mathbb{R}^{2K \times d_p}$ are then reshaped and recombined to form the reconstructed complex channel matrix $\hat{\mathbf{H}} \in \mathbb{C}^{N_{\text{s}} \times N_{\text{f}}}$.

\subsection{\reviewer{Contrastive Head}}

% Building upon the WiMAE architecture described above, we implement a contrastive learning extension that 

\reviewer{Contrastive head} operates in parallel with the reconstruction pathway. This section details the specific implementation aspects of the contrastive component.

\subsubsection*{Global feature extraction}

To obtain a compact representation of the encoder output $\mathbf{Z}_{\text{enc}} \in \mathbb{R}^{2N_{\text{v}} \times d_{\text{e}}}$, we first apply mean pooling across all visible patches to calculate $\mathbf{z}_{\text{mean}} \in \mathbb{R}^{d_{\text{e}}}$:
\begin{equation}
\mathbf{z}_{\text{mean}} = \frac{1}{2N_{\text{v}}} \sum_{i=1}^{2N_{\text{v}}} \mathbf{Z}_{\text{enc}}[i, :],
\label{eq:mean_pool}
\end{equation}
where $\mathbf{Z}_{\text{enc}}[i, :]$ is the $i$-th row of $\mathbf{Z}_{\text{enc}}$.

\subsubsection*{Projection to contrastive embeddings}

$\mathbf{z}_{\text{mean}}$ is projected to the contrastive embedding space with a two-layer MLP with a ReLU activation function, i.e.,
\begin{equation}
\text{MLP}^{\text{C}}(\mathbf{z}_{\text{mean}}) = \mathbf{W}^{\text{C}}_2 \text{ReLU}(\mathbf{W}^{\text{C}}_1 \mathbf{z}_{\text{mean}} + \mathbf{b}^{\text{C}}_1) + \mathbf{b}^{\text{C}}_2,
\label{eq:mlp_contra}
\end{equation}
where $\mathbf{W}^{\text{C}}_1 \in \mathbb{R}^{d_{\text{e}} \times 2d_{\text{c}}}$ and $\mathbf{W}^{\text{C}}_2 \in \mathbb{R}^{2d_{\text{c}} \times d_{\text{c}}}$ are weight matrices, $\mathbf{b}^{\text{C}}_1 \in \mathbb{R}^{2d_{\text{c}}}$ and $\mathbf{b}^{\text{C}}_2 \in \mathbb{R}^{d_{\text{c}}}$ are bias terms, and $d_{\text{c}}$ is the dimension of the contrastive embedding space. 
The final contrastive embedding $\mathbf{z}_{\text{c}} \in \mathbb{R}^{d_{\text{c}}}$ is obtained by applying L2 normalization to the projection output:
\begin{equation}
\mathbf{z}_{\text{c}} = \frac{\text{MLP}^{\text{C}}(\mathbf{z}_{\text{mean}})}{\|\text{MLP}^{\text{C}}(\mathbf{z}_{\text{mean}})\|_2} \in \mathbb{R}^{d_{\text{c}}}.
\label{eq:l2_norm}
\end{equation}

\section{Methodology}\label{sec:sim}
This section details the methodology used in our experimental evaluation of ContraWiMAE, including dataset generation, pretraining setup, and adaptations for downstream tasks. Through these experiments, we aim to demonstrate the effectiveness of our self-supervised learning approach for wireless channel representation learning. A thorough discussion of the results is provided in Section~\ref{sec:results}.

\subsection{Dataset Generation} \label{sec:data_gen}
\reviewer{For our experimental evaluation, we configure the system with $N_{\text{f}}=32$ subcarriers, $\Delta f = 30$\,kHz subcarrier spacing (total bandwidth $B=960$\,kHz), and $N_\text{s}=32$ antennas with half-wavelength spacing. All antennas have isotropic radiation patterns and single polarization, with a maximum of 20 propagation paths considered.} Using the DeepMIMO dataset \cite{deepmimo}, we generate \reviewer{2.5}M samples at 3.5\,GHz operating frequency from \reviewer{56} scenarios for pretraining. Each sample in the dataset corresponds to a channel for a distinct user-base station \reviewer{pair} from one of the following scenarios: \reviewer{ \textit{Amsterdam},} \textit{ASU Campus}, \reviewer{\textit{Athens}, \textit{Bangkok}, \textit{Barcelona}, \textit{Beijing},} \textit{Boston}, \reviewer{\textit{Brussels}, \textit{Cairo}, \textit{Cape Town}, \textit{Charlotte},} \textit{Chicago}, \reviewer{ \textit{Denver}, \textit{Dubai}, \textit{Edinburgh}, \textit{Florence}, \textit{Fort Worth}, \textit{Fujiyoshida}, \textit{Granada}, \textit{Hatsukaichi}, \textit{Helsinki}, \textit{Hong Kong}, \textit{Indianapolis}, \textit{Istanbul}, \textit{Jerusalem}, \textit{Kyoto}, \textit{Havana}, \textit{Lisbon},} \textit{Los Angeles}, \reviewer{ \textit{Madrid}, \textit{Marrakesh}, \textit{Mumbai}, \textit{New Delhi},} \textit{New York}, \reviewer{ \textit{North Jakarta}, \textit{Oklahoma City},} \textit{Philadelphia}, \textit{Phoenix}, \reviewer{ \textit{Reykjavik}, \textit{Rio de Janeiro}, \textit{Rome},} \textit{San Francisco}, \reviewer{ \textit{San Nicolas}, \textit{Saint Petersburg}, \textit{Santa Clara}, \textit{Santiago},} \textit{Seattle}, \reviewer{ \textit{Seoul}, \textit{Singapore}, \textit{Stockholm}, \textit{Sumida City}, \textit{Sydney}, \textit{Taipei}, \textit{Taito City}, \textit{Toronto}, and \textit{Warsaw}}. We also generate a dataset for downstream tasks with \reviewer{0.55M total} samples from \reviewer{10} unseen scenarios: \reviewer{\textit{Austin}, \textit{Centro}, \textit{Columbus}, \textit{Dallas}, \textit{Gurbchen}, \textit{Houston}, \textit{Miami}, \textit{Montreal}, \textit{Prague}, and} \textit{San Diego}. Similar to the pretraining dataset, \reviewer{some} scenarios feature \reviewer{more than} one base station with uniformly distributed users. \reviewer{Apart from the 3.5\,GHz channels, we generate channels at 28\,GHz for downstream task scenarios with available ray-tracing data. To ensure fairness in comparisons involving LWM, we also use a smaller set of scenarios exactly matching the configuration described in the respective work \cite{lwm} for both the pretraining and downstream datasets. This dataset consists of 1.14M samples for pretraining and 14,840 samples for downstream tasks.}

\reviewer{We selected the DeepMIMO dataset for three key reasons: (i) to enable direct benchmarking with prior work such as LWM \cite{lwm} under identical input features and data-generation pipelines, (ii) to ensure reproducibility by providing a publicly available dataset that enables other researchers to retrain and validate our approach, and (iii) to test generalization across diverse geographical layouts, leveraging DeepMIMO's extensive collection of over 50 ray-traced scenarios spanning cities worldwide.}

\subsection{\reviewer{Reconstruction-only Pretraining}}
\label{sec:meth:pretraining_of_wimae}
 \reviewer{For brevity, we refer to our reconstruction-only foundation model as WiMAE in the following sections.} We train WiMAE for 3,000 epochs with a \reviewer{90/10} training/validation split \reviewer{and a batch size of 8,096}. We use the \reviewer{AdamW} optimizer with \reviewer{linear rate warm-up reaching $3 \times 10^{-4}$ in the first 10 epochs, followed by cosine} decay to $3 \times 10^{-6}$.

\reviewer{While we explore many different configurations, as discussed in detail in Section \ref{sec:results},}
for the comparisons with baseline models on the downstream tasks, we set $m_{\text{r}}=\reviewer{0.90}$, $L_{\text{enc}}=12$, $L_{\text{dec}}=4$, $d_{\text{e}}=64$, \reviewer{and $(N_{\text{p,s}}, N_{\text{p,f}}) = \reviewer{(16, 1)}$},  \reviewer{striking a balance between complexity and performance}. These choices result in 31\% fewer encoder parameters and 7\% fewer total parameters compared with LWM (see Table~\ref{tab:params}). \reviewer{Here, we note that the choice of $m_\text{r}$ does not affect the number of parameters, and hence the model expressivity. However, it directly affects the pretext task difficulty, necessitating searching for the optimal mask ratio for each patch shape for optimal learning.} For LWM, we use publicly available pretrained weights and the code to generate its embeddings \cite{lwm} without retraining or modifications.

\subsection{Pretraining of ContraWiMAE}

We warm-start the encoder and decoder components of ContraWiMAE 
% by copying the corresponding weights
from \reviewer{WiMAE.} 
% to preserve its reconstruction capability.
We use the same learning rate schedule, optimizer, and batch size for continual pretraining of ContraWiMAE on the same pretraining dataset. To maximize the transferability of the reconstruction capability, we use %$m_{\text{r}}=\reviewer{0.90}$, $L_{\text{enc}}=12$, $L_{\text{dec}}=4$, $d_{\text{e}}=64$, and $(N_{\text{p,s}}, N_{\text{p,f}}) = \reviewer{(16, 1)}$, matching 
\reviewer{the same parameters as those used in the reconstruction foundation.} 
% for WiMAE used in downstream task evaluations. 

We set the \reviewer{reconstructive-contrastive trade-off weight} $\alpha=0.9$ and the temperature parameter \reviewer{$\tau=0.2$}. We also set the contrastive space dimension to $d_{\text{c}}=64$ and uniformly sample the noise variance for positive pairs to achieve channel SNR values between \reviewer{5}\,dB and \reviewer{40}\,dB. \reviewer{We choose these contrastive learning parameters through ablation studies based on the total loss in Eq.~\eqref{eqn:combined_objective}.}

\begin{table}[!t]
\centering
\renewcommand{\arraystretch}{1.2}
\caption{Parameter counts of different configurations.}
\small
\begin{tabular}{lcccc}
\hline
\textbf{Model} & \textbf{Encoder} & \textbf{Decoder} & \textbf{C. Head} & \textbf{Total} \\
\hline
CWiMAE (12,4) & 410,944 & 143,184 & 16,576 & 570,704 \\
WiMAE (6,4) & 210,112 & 143,184 & - & 353,296 \\
WiMAE (12,4) & 410,944 & 143,184 & - & 554,128 \\
WiMAE (18,4) & 611,776 & 143,184 & - & 754,960 \\
WiMAE (12,8) & 410,944 & 277,072 & - & 688,016 \\
WiMAE (12,12) & 410,944 & 410,960 & - & 821,904 \\
LWM (12,-) & 600,000 & - & - & 600,000 \\
\hline
\end{tabular}
\label{tab:params}
\vspace{0.5mm}

\footnotesize{\textit{Note: CWiMAE = ContraWiMAE. Model names show ($L_{\text{enc}}$,$L_{\text{dec}}$) where $L_{\text{enc}}$ and $L_{\text{dec}}$ are encoder and decoder layers. For WiMAE and CWiMAE: $M_{\text{enc}}{=}16$, $M_{\text{dec}}{=}8$, $d_{\text{e}}{=}64$.}}
\end{table}

\subsection{Downstream Task Adaptation} \label{sec:downtask}
We evaluate ContraWiMAE on \reviewer{three} downstream tasks: (i) cross-frequency beam selection, (ii) line-of-sight (LoS) detection, and \reviewer{(iii) channel estimation.} \reviewer{For LoS detection and beam selection, we perform both parametric and non-parametric evaluations where we freeze the pretrained encoder. For parametric evaluation, which adds expressivity to the prediction pipeline, we train a linear classifier on top of the encoder representations, following standard practices. We also train custom residual networks to see the marginal improvement possible through learning task-specific non-linearities. For our non-parametric evaluation, we assess the performance of a kNN classifier on the pretrained encoder representations, disentangling representation quality from task head capacity. For channel estimation, we keep the encoder-decoder pair and measure the training-free performance. We also analyze the gain from fine-tuning the model on this specific task.} 
\begin{table*}[t]
\centering
\caption{Top-1/top-3 accuracy for beam selection with linear probing.}
\label{tab:beam_selection}
\setlength{\tabcolsep}{8pt}
\renewcommand{\arraystretch}{1.2}
\begin{tabular}{|c|c||c|c|c|c|c|c|c|}
\hline
\textbf{Model} & \textbf{Codebook Size (CS)} & \multicolumn{7}{c|}{\textbf{Training Budget}} \\
\cline{3-9}
& & \textbf{1\%} & \textbf{2\%} & \textbf{5\%} & \textbf{10\%} & \textbf{25\%} & \textbf{50\%} & \textbf{100\%} \\
\hline\hline
\multirow{5}{2cm}{\centering\textbf{WiMAE}} 
& 16 & $22.6$/$36.1$ & $32.2$/$48.5$ & $27$/$42.7$ & $31.7$/$43.5$ & $36.1$/$50.2$ & $38.9$/$49.5$ & $41.4$/$51$ \\
\cline{2-9}
& 32 & $15.6$/$28.3$ & $18.9$/$26.4$ & $19.8$/$28$ & $23.8$/$32.8$ & $28.2$/$34$ & $33$/$37.2$ & $31.5$/$36.6$ \\
\cline{2-9}
& 64 & $13$/$19.5$ & $14.4$/$23.1$ & $13$/$18.3$ & $17.8$/$22.6$ & $24.1$/$29.8$ & $26.3$/$31.7$ & $26.6$/$30.4$ \\
\cline{2-9}
& 128 & $7.2$/$10.3$ & $9.6$/$15.1$ & $10.2$/$15.7$ & $14.2$/$18.7$ & $18.9$/$22$ & $19.1$/$23.4$ & $16.6$/$19.1$ \\
\cline{2-9}
& 256 & $5$/$8.4$ & $6.7$/$11.5$ & $6.9$/$11.2$ & $9.4$/$13.7$ & $10.5$/$13.7$ & $12.3$/$14.8$ & $10.2$/$11.8$ \\
\hline\hline
\multirow{5}{2cm}{\centering\textbf{ContraWiMAE}} 
& 16 & $\textbf{49.2}$/$\textbf{69.7}$ & $\textbf{55.9}$/$\textbf{75.7}$ & $\textbf{62.7}$/$\textbf{79.4}$ & $\textbf{72.1}$/$\textbf{87.2}$ & $\textbf{72.2}$/$\textbf{83.6}$ & $\textbf{77.4}$/$\textbf{87.5}$ & $\textbf{79}$/$\textbf{89.5}$ \\
\cline{2-9}
& 32 & $\textbf{35.9}$/$\textbf{47.4}$ & $\textbf{41.4}$/$\textbf{57.1}$ & $\textbf{50.6}$/$\textbf{68.2}$ & $\textbf{63.5}$/$\textbf{77.5}$ & $\textbf{68.4}$/$\textbf{76.3}$ & $\textbf{68.3}$/$\textbf{76.6}$ & $\textbf{75.3}$/$\textbf{84.2}$ \\
\cline{2-9}
& 64 & $\textbf{24.7}$/$\textbf{36.2}$ & $\textbf{29}$/$\textbf{44.6}$ & $\textbf{31.3}$/$\textbf{46.3}$ & $\textbf{40.1}$/$\textbf{54.3}$ & $\textbf{48.3}$/$\textbf{62.8}$ & $\textbf{56.1}$/$\textbf{69.7}$ & $\textbf{64.5}$/$\textbf{79}$ \\
\cline{2-9}
& 128 & $\textbf{14}$/$\textbf{20.6}$ & $\textbf{16.2}$/$\textbf{27.7}$ & $\textbf{18.5}$/$\textbf{30.3}$ & $\textbf{23.7}$/$\textbf{34.7}$ & $\textbf{29}$/$\textbf{39.5}$ & $\textbf{35.4}$/$\textbf{46.3}$ & $\textbf{39.3}$/$\textbf{49.6}$ \\
\cline{2-9}
& 256 & $\textbf{8}$/$\textbf{14.5}$ & $\textbf{10.9}$/$\textbf{18.7}$ & $11.8$/$\textbf{21.5}$ & $\textbf{15.2}$/$23.3$ & $\textbf{18}$/$26.6$ & $\textbf{18.9}$/$26.1$ & $20.2$/$26.6$ \\
\hline\hline
\multirow{5}{2cm}{\centering\textbf{LWM}} 
& 16 & $16.3$/$31.5$ & $22.4$/$33.7$ & $18.4$/$33$ & $25.3$/$36.1$ & $36.7$/$46.2$ & $37$/$47$ & $44.1$/$54.6$ \\
\cline{2-9}
& 32 & $16.4$/$25.3$ & $20.1$/$27.2$ & $14.2$/$22.6$ & $25.7$/$34.9$ & $36$/$44.3$ & $37.1$/$45.1$ & $39.6$/$46.8$ \\
\cline{2-9}
& 64 & $14.8$/$21$ & $18.5$/$25.9$ & $12.1$/$17.5$ & $20$/$26.6$ & $26.2$/$33.3$ & $32.4$/$39$ & $35.4$/$41.5$ \\
\cline{2-9}
& 128 & $9$/$14$ & $10.7$/$18.4$ & $7.6$/$12.7$ & $12.3$/$19.5$ & $17.5$/$23.3$ & $21.6$/$27.8$ & $23.7$/$29.4$ \\
\cline{2-9}
& 256 & $6$/$9.5$ & $7$/$11.8$ & $5.3$/$9.1$ & $7.6$/$11.7$ & $11.1$/$16.2$ & $12.4$/$18$ & $12.5$/$17.7$ \\
\hline\hline
\multirow{5}{2cm}{\centering\textbf{RAW}} 
& 16 & $31.7$/$40.3$ & $31.7$/$41.8$ & $30.8$/$41.6$ & $38.1$/$46.2$ & $39.1$/$47.1$ & $38.9$/$44.9$ & $41.2$/$47.4$ \\
\cline{2-9}
& 32 & $27.6$/$38.5$ & $24.5$/$36.7$ & $34.1$/$43.3$ & $36.8$/$46.4$ & $36.5$/$43.8$ & $41.2$/$47$ & $42.5$/$48.1$ \\
\cline{2-9}
& 64 & $19.4$/$31.1$ & $22.6$/$34.3$ & $27.6$/$40.8$ & $30.5$/$42.4$ & $33.5$/$43.5$ & $33.6$/$42.6$ & $36.1$/$43.7$ \\
\cline{2-9}
& 128 & $9.9$/$18.2$ & $12.7$/$23.4$ & $17.4$/$29.5$ & $19.9$/$31.8$ & $23.9$/$35$ & $27$/$38.5$ & $29.3$/$37.8$ \\
\cline{2-9}
& 256 & $5.9$/$11.7$ & $9$/$16$ & $\textbf{12}$/$20.1$ & $13.5$/$\textbf{23.8}$ & $16.6$/$\textbf{29}$ & $17.1$/$\textbf{29.6}$ & $\textbf{20.9}$/$\textbf{31.6}$ \\
\hline
\end{tabular}
\label{tab:linear_beam_selection}
\vspace{1mm}
\footnotesize{\textit{Note: \reviewer{RAW refers to models trained directly on raw channel data without any representation learning.} Boldface indicates the best result for each (CS, training budget) pair.}}
\end{table*}

\reviewer{These tasks probe complementary aspects of the learned representations.} The beam selection task evaluates how well each encoding captures the channel's detailed site-specific structure across antenna elements and subcarriers, while the LoS \reviewer{detection} measures the effectiveness of the extracted features as compact channel summaries. \reviewer{Channel estimation assesses reconstruction capability by evaluating the model's ability to recover the underlying channel from sparse and noisy observations.}

\subsubsection{Cross-frequency Beam Selection}
The beam selection task predicts optimal mmWave beams at 28\,GHz from 3.5\,GHz channel data using predefined codebooks with uniform beam patterns and codebook sizes (CS) of 16, 32, 64, 128, and 256. This reduces the channel estimation overhead that is normally needed in beam management \cite{GioPolRoy2019} while testing how well the learned representations capture cross-frequency propagation characteristics. We report top-1 and top-3 accuracy: top-1 accuracy measures the proportion of test samples for which the predicted best beam matches the true optimal beam (the beam that maximizes the received power); top-3 accuracy extends this by considering a prediction correct if the true optimal beam is among the top three highest-scoring predicted beams. We compare models trained on ContraWiMAE embeddings with models trained on LWM\reviewer{, WiMAE, and supervised} embeddings and models trained on raw channel data without any representation learning (RAW), as summarized in Tables~\ref{tab:linear_beam_selection} and~\ref{tab:resnet_beam_selection} \reviewer{and several figures in Section~\ref{sec:results}. The supervised baseline consists of an encoder (same as ContraWiMAE) and a linear classifier attached to it. However, we train this model directly on the pretraining dataset with 3.5\,GHz channel-beam pairs separately for each codebook size. We discard the linear classifier and use the encoder representations for downstream tasks.}

\subsubsection*{Downstream models}
We employ several models for downstream evaluation.
\begin{itemize}
\item
\emph{Linear Probing}: This approach evaluates latent space linear separability by freezing encoder weights and training a linear classifier on top of the encoder to analyze how information can be directly extracted from the features without introducing nonlinearities \cite{AlaBen2018}. As a lightweight and computationally efficient method, linear probing offers faster training and inference. To train a linear classifier, we minimize the cross-entropy loss using Adam optimizer with initial learning rate of $1 \times 10^{-4}$, exponential decay with $\gamma=0.995$, and batch size of 512 until convergence.

\item
\emph{ResNet Models}: We implement two residual network variants \cite{He_2016_CVPR}. \emph{ResNet-Wide}, the larger architecture with 57,000 parameters, features deeper layers with wider channel dimensions starting at 24 and expanding to 96 with a moderate dropout of 0.4. \emph{ResNet-Slim}, the compact variant with 27,000 parameters, uses narrower channel dimensions starting at 16 and expanding to 64, reduced block counts and a lighter dropout of 0.3. Both architectures process encoded features through 1D convolutional layers with residual connections and batch normalization before classification with a fully connected network. For ResNet models, we use the same optimization parameters as linear probing since additional hyperparameter tuning does not lead to notable performance improvement.

\color{black}
\item
\emph{kNN Classifier}: We adopt a sampling-based evaluation protocol where multiple random subsets are drawn from both the training and test scenarios from the downstream dataset. We report the mean accuracy results considering $k=10$ nearest neighbors in Euclidean space. Our sampling approach aims to mitigate selection bias, providing robust performance estimates.
\color{black}
\end{itemize}

\begin{table*}[t]
\centering
\caption{Top-1/top-3 accuracy for beam selection with ResNet-Slim and ResNet-Wide.}
\label{tab:beam_selection_comparison}
\setlength{\tabcolsep}{6pt}
\renewcommand{\arraystretch}{1.2}
\begin{tabular}{|c|c||c|c|c|c||c|c|c|c|}
\hline
\multirow{3}{*}{\textbf{Model}} & \multirow{3}{*}{\textbf{Codebook Size (CS)}} & \multicolumn{8}{c|}{\textbf{Training Budget}} \\
\cline{3-10}
& & \multicolumn{4}{c||}{\textbf{ResNet-Slim}} & \multicolumn{4}{c|}{\textbf{ResNet-Wide}} \\
\cline{3-10}
& & \textbf{10\%} & \textbf{25\%} & \textbf{50\%} & \textbf{100\%} & \textbf{10\%} & \textbf{25\%} & \textbf{50\%} & \textbf{100\%} \\
\hline\hline
\multirow{5}{1.75cm}{\centering\textbf{WiMAE}} 
& 16 & $8.9/23.8$ & $37.8/71.6$ & $42.9/74.7$ & $53.7/82.2$ & $6.9/24.7$ & $55.6/84.8$ & $61.3/86.1$ & $68.9/89.2$ \\
\cline{2-10}
& 32 & $13.4/27.1$ & $25.7/47.5$ & $39.2/70.1$ & $52/79.6$ & $28.4/51.5$ & $47.3/76.5$ & $60.2/86.9$ & $71.1/92.1$ \\
\cline{2-10}
& 64 & $2.2/6.1$ & $11.1/27.8$ & $24.1/49$ & $36.8/65.4$ & $14.7/30.3$ & $32.2/59.3$ & $47.6/77.2$ & $60.7/86.7$ \\
\cline{2-10}
& 128 & $0.7/2.8$ & $6.4/13.7$ & $12.7/30.7$ & $24.4/50.3$ & $9.2/18.1$ & $17.5/35.8$ & $32.6/61.9$ & $44.3/76.1$ \\
\cline{2-10}
& 256 & $0.2/1.6$ & $0.4/1.3$ & $8.1/17.5$ & $13.1/29.4$ & $5.5/10$ & $8.8/17.9$ & $15/33.9$ & $22.2/51.2$ \\
\hline\hline
\multirow{5}{1.75cm}{\centering\textbf{ContraWiMAE}} 
& 16 & $\textbf{31.2}$/$\textbf{55}$ & $\textbf{56.7}$/$\textbf{80.7}$ & $\textbf{68.9}$/$\textbf{88.8}$ & $\textbf{73.9}$/$\textbf{92.3}$ & $\underline{\textbf{65.5}}$/$\underline{\textbf{87.6}}$ & $\underline{\textbf{74.6}}$/$\underline{\textbf{91.9}}$ & $\underline{\textbf{77}}$/$\underline{\textbf{93.7}}$ & $\underline{\textbf{79}}$/$\underline{\textbf{93.7}}$ \\
\cline{2-10}
& 32 & $\textbf{15.7}$/$\textbf{33.6}$ & $\textbf{46.8}$/$\textbf{76.8}$ & $\textbf{64.7}$/$\textbf{88.8}$ & $\textbf{73.8}$/$\textbf{93.3}$ & $\underline{\textbf{49.3}}$/$\underline{\textbf{75.3}}$ & $\underline{\textbf{72}}$/$\underline{\textbf{93}}$ & $\underline{\textbf{79.1}}$/$\underline{\textbf{95}}$ & $\underline{\textbf{81.9}}$/$\underline{\textbf{95.8}}$ \\
\cline{2-10}
& 64 & $\textbf{10.4}$/$\textbf{22.6}$ & $\textbf{24.2}$/$\textbf{49.2}$ & $\textbf{40.4}$/$\textbf{70.8}$ & $\textbf{63}$/$\textbf{88.9}$ & $\underline{\textbf{26.4}}$/$\underline{\textbf{51.4}}$ & $\underline{\textbf{55.5}}$/$\underline{\textbf{82.9}}$ & $\underline{\textbf{69.2}}$/$\underline{\textbf{93.1}}$ & $70.7$/$\underline{\textbf{93.5}}$ \\
\cline{2-10}
& 128 & $\textbf{3.2}$/$\textbf{8}$ & $\textbf{11.6}$/$\textbf{29.9}$ & $\textbf{27.3}$/$\textbf{57.5}$ & $\textbf{46.6}$/$\textbf{78.9}$ & $\underline{\textbf{15}}$/$\underline{\textbf{30.8}}$ & $\underline{\textbf{31.9}}$/$\underline{\textbf{57.4}}$ & $\underline{\textbf{50.4}}$/$\underline{\textbf{82.1}}$ & $\underline{\textbf{59.6}}$/$\underline{\textbf{88.5}}$ \\
\cline{2-10}
& 256 & $\textbf{2.9}$/$\textbf{5.7}$ & $\textbf{6.6}$/$\textbf{14.9}$ & $\textbf{12.7}$/$\textbf{29.2}$ & $\textbf{24.5}$/$\textbf{56.3}$ & $\underline{\textbf{7.6}}$/$\underline{\textbf{13.8}}$ & $\underline{\textbf{12.4}}$/$\underline{\textbf{29.8}}$ & $\underline{\textbf{24.9}}$/$\underline{\textbf{53.1}}$ & $\underline{\textbf{39.4}}$/$\underline{\textbf{74.7}}$ \\
\hline\hline
\multirow{5}{1.75cm}{\centering\textbf{LWM}} 
& 16 & $23.3/50.9$ & $49.3/79.1$ & $55.1/85$ & $64.4/88.8$ & $49.2/80.5$ & $65.6/90.1$ & $71.5/92.6$ & $77.3/93.3$ \\
\cline{2-10}
& 32 & $12/30$ & $35/66.2$ & $48.7/79.9$ & $69.8/92.9$ & $33.1/59.6$ & $58.4/86.2$ & $70.7/93.6$ & $77/95.4$ \\
\cline{2-10}
& 64 & $2/5$ & $15.7/35.4$ & $29.3/62.1$ & $51.6/84.1$ & $17.9/38.7$ & $39.1/72.1$ & $55.1/87.2$ & $64.2/92.6$ \\
\cline{2-10}
& 128 & $1.1/3.3$ & $9/20.8$ & $11.8/32.6$ & $28.4/65.2$ & $8/17.8$ & $14.3/35.4$ & $30.9/67.6$ & $47.2/81.4$ \\
\cline{2-10}
& 256 & $0.4/1.5$ & $4.5/10.6$ & $7.1/16.3$ & $11.9/35.7$ & $4.7/9.4$ & $8/18.9$ & $14.2/32.6$ & $24.3/55.3$ \\
\hline\hline
\multirow{5}{1.75cm}{\centering\textbf{RAW}} 
& 16 & $9.5/22$ & $29.4/55.7$ & $57.7/80.3$ & $70/88$ & $28/52.4$ & $47.4/74.6$ & $69/88.4$ & $75.9/91.1$ \\
\cline{2-10}
& 32 & $3.4/12.1$ & $26.9/49.1$ & $43.3/72.2$ & $63.1/86.8$ & $23.9/41$ & $46/74.3$ & $67.5/88.6$ & $80.3/94.3$ \\
\cline{2-10}
& 64 & $2.6/7.4$ & $15.6/30.7$ & $31.4/53.1$ & $57.4/84.9$ & $13.8/27$ & $35.4/58.8$ & $51.1/79.6$ & $\underline{\textbf{73.4}}$/$92.4$ \\
\cline{2-10}
& 128 & $1.7/4.2$ & $2.3/5.7$ & $16.3/36.6$ & $34.6/62.1$ & $1/3.4$ & $22.3/43.4$ & $37.5/62.5$ & $55.9/81.5$ \\
\cline{2-10}
& 256 & $1/2.4$ & $0.8/1.8$ & $10.4/23.8$ & $18.7/41.5$ & $0.7/3.4$ & $10.1/21.2$ & $21.3/44.1$ & $38.3/69.3$ \\
\hline
\end{tabular}
\vspace{1mm}
\footnotesize{\textit{Note: Best results for slim and wide are bolded and both bolded and underlined, respectively.}}
\label{tab:resnet_beam_selection}
\end{table*}

\subsubsection{LoS Detection}
This task classifies if a channel exhibits the LoS condition, which can be crucial for adaptive modulation and coding, blockage prediction, and predictive handover. We report accuracy, F1 score, and area under the ROC curve (AUC), comparing the ContraWiMAE representations with the \reviewer{WiMAE} and LWM baselines. In our comparison, we only employ linear probing since complex downstream models offer minimal additional improvement for this task. We use the CLS token for LWM and mean-pooled embeddings for WiMAE and ContraWiMAE. Although WiMAE and ContraWiMAE do not have a classification token like CLS, calculating the mean value across each patch encoding serves as a summary of the input. All experiments, shown in Table~\ref{tab:linear_los}, use Adam optimizer with a learning rate of 0.01 with exponential decay with $\gamma=0.995$ and a batch size of 256.

\begin{table*}[t]
\centering
\caption{Accuracy in the LoS detection task.}
\setlength{\tabcolsep}{11pt}
\renewcommand{\arraystretch}{1.2}
\begin{tabular}{|c||c|c|c||c|c|c||c|c|c|}
\hline
\multirow{2}{*}{\textbf{Training Budget (\%)}} & \multicolumn{3}{c||}{\textbf{LWM}} & \multicolumn{3}{c||}{\textbf{WiMAE}} & \multicolumn{3}{c|}{\textbf{ContraWiMAE}} \\
\cline{2-10}
& Accuracy & F1 & AUC & Accuracy & F1 & AUC & Accuracy & F1 & AUC \\
\hline\hline
1 & $0.929$ & $0.941$ & $0.981$ & $\textbf{0.945}$ & $\textbf{0.956}$ & $\textbf{0.990}$ & $0.937$ & $0.949$ & $0.984$ \\
\hline
2 & $0.946$ & $0.957$ & $0.985$ & $\textbf{0.947}$ & $0.957$ & $\textbf{0.989}$ & $\textbf{0.947}$ & $\textbf{0.958}$ & $0.986$ \\
\hline
5 & $0.926$ & $0.942$ & $0.973$ & $\textbf{0.946}$ & $\textbf{0.957}$ & $\textbf{0.990}$ & $0.945$ & $0.956$ & $0.986$ \\
\hline
10 & $0.943$ & $0.955$ & $0.987$ & $\textbf{0.951}$ & $\textbf{0.960}$ & $\textbf{0.990}$ & $0.945$ & $0.956$ & $0.989$ \\
\hline
25 & $0.946$ & $0.957$ & $0.988$ & $\textbf{0.957}$ & $\textbf{0.965}$ & $\textbf{0.992}$ & $0.950$ & $0.960$ & $0.990$ \\
\hline
50 & $0.950$ & $0.960$ & $0.989$ & $\textbf{0.958}$ & $\textbf{0.966}$ & $\textbf{0.992}$ & $0.951$ & $0.961$ & $0.991$ \\
\hline
100 & $0.948$ & $0.959$ & $0.989$ & $\textbf{0.960}$ & $\textbf{0.968}$ & $\textbf{0.993}$ & $0.953$ & $0.962$ & $0.991$ \\
\hline
\end{tabular}
\vspace{1mm}
\footnotesize{\textit{Note: Boldface indicates best result for each training budget.}}
\label{tab:linear_los}
\end{table*}

\begin{figure}[!h]
    \centering
    \begin{subfigure}{0.49\textwidth}
        \centering
        \includegraphics[width=\textwidth]{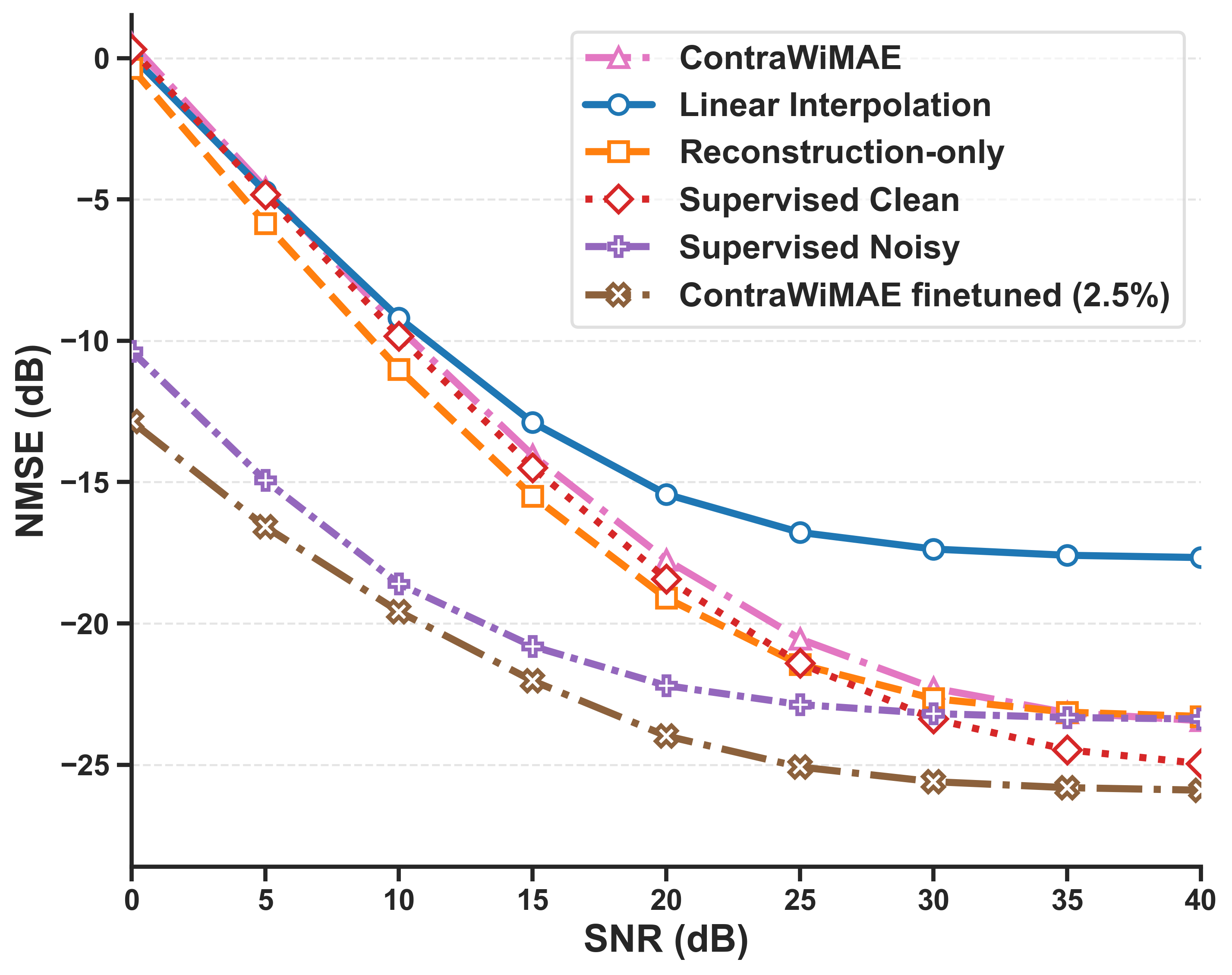}
        \caption{ContraWiMAE vs. baselines}
        \label{fig:ce}
    \end{subfigure}
    \hfill
    \begin{subfigure}{0.49\textwidth}
        \centering
        \includegraphics[width=\textwidth]{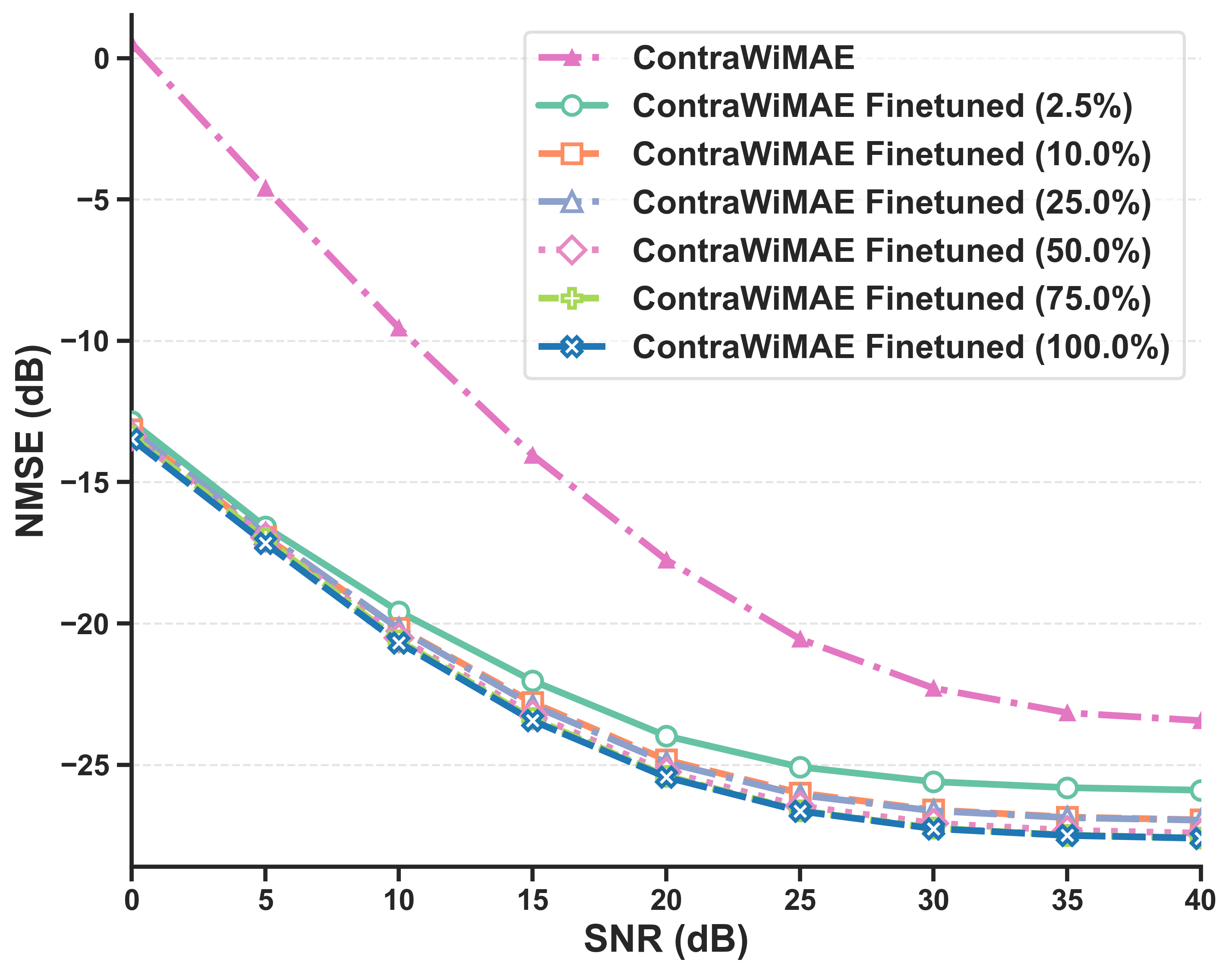}
        \caption{ContraWiMAE finetuning gain}
        \label{fig:ce_ft}
    \end{subfigure}
    
    \caption{Channel estimation on unseen scenarios}
    \label{fig:ce_main}
\end{figure}
\color{black}
\subsubsection{Channel Estimation}
We evaluate an uplink configuration where the user equipment transmits a pilot tone on a single subcarrier. The base station captures channel measurements at each antenna element, with additive Gaussian noise corrupting these observations. Using the pretrained ContraWiMAE, the base station reconstructs the full channel from this partial, noisy measurement. We keep the pretrained decoder for this task along with the encoder, eliminating the need to train a separate decoder. We assess performance using normalized mean square error (NMSE) on held-out test data (\textit{Miami} and \textit{Houston} scenarios) from the downstream task dataset, with test SNR values spanning from 0 to 40\,dB at 5\,dB intervals. Note that in this setting, we do not finetune or adapt ContraWiMAE for the specific pilot pattern on the selected scenarios. Therefore, to further investigate the performance gain through finetuning on the selected pilot pattern, we utilize the remaining scenarios from the downstream task dataset with finetuning budgets set at 2.5\%, 10\%, 25\%, 50\%, 75\%, and 100\% (see Fig.~\ref{fig:ce_ft}), where 100\% indicates utilizing all remaining task scenarios for adaptation.

In Section \ref{sec:results}, we benchmark against three baselines: (1) a supervised encoder-decoder pair, sharing the same architectural details as the ContraWiMAE encoder and decoder, trained on the pretraining dataset with the same pilot pattern as the test, (2) a supervised encoder-decoder pair as explained above but trained with noise-corrupted pretraining data matching the test SNR levels, and (3) a linear interpolation method.
\color{black}

\color{black}  % color the entire section red
\section{Computational Complexity Analysis}
\label{sec:complexity}

This section provides a detailed analysis of the computational complexity of ContraWiMAE. We examine the forward time complexity during both pretraining and inference phases, providing insights into how these scale with respect to key architectural parameters.

\subsection{Preprocessing Complexity}
% \textit{Partitioning and masking:} 
The partitioning operation segments $\mathbf{H}$ into $K$ patches and applies a binary mask at the patch level, requiring $\mathcal{O}(K)$ operations. Separating each visible patch into real and imaginary components requires $\mathcal{O}(N_\text{v} d_\text{p})$ operations, where each complex value is decomposed into two real values. Flattening visible patches in a sequence form as in Eq.~\eqref{eq:vectorization} requires $\mathcal{O}(2N_\text{v} d_\text{p})$ operations to rearrange memory for both real and imaginary components. The projection in Eq.~\eqref{eq:linear_proj} for the visible patch sequence % $\mathbf{P}_\text{v} \in \mathbb{R}^{2N_\text{v} \times d_\text{p}}$ with $\mathbf{W}_\text{0} \in \mathbb{R}^{d_\text{p} \times d_\text{e}}$ 
requires $\mathcal{O}(2N_\text{v} d_\text{p} d_\text{e})$. Adding positional encodings, as in Eq.~\eqref{eq:pos_enc}, is $\mathcal{O}(2N_\text{v} d_\text{e})$, which is negligible compared to the projection. Therefore, the total preprocessing complexity $\mathcal{C}_{\text{prep}}$, dominated by the linear projection, is $\mathcal{O}(2N_\text{v}Kd_\text{p} d_\text{e})$.

\begin{table*}[t]
\centering
\caption{Run-time measurements}
\begin{tabular}{lccccc}
\toprule
\textbf{Metric} & \textbf{Encoder-only} & \textbf{Encoder-decoder } & \textbf{Encoder-decoder } & \textbf{Encoder-decoder } & \textbf{Encoder-decoder } \\
 & $m_\text{r}=0$ & $m_\text{r}=0.45$ & $m_\text{r}=0.60$ & $m_\text{r}=0.75$ & $m_\text{r}=0.90$ \\
\midrule
Latency (ms) & 0.237 & 0.157 & 0.127 & 0.099 & 0.057 \\
Throughput (samples/s) & 4221.2 & 6371.9 & 7875.7 & 10085.2 & 17534.1 \\
Peak Memory (MB) & 1.156 & 0.440 & 0.270 & 0.236 & 0.231 \\
GFLOPs & 0.126 & 0.082 & 0.063 & 0.048 & 0.033 \\
\bottomrule
\end{tabular}
\label{tab:inference_performance}
\vspace{0.5mm}

{\centering\footnotesize\textit{Note: We report per-sample average values over 100 forward passes with batch size of 128 on NVIDIA RTX A4000 (PyTorch 2.5, CUDA 12.1)}\par}
\end{table*}

\subsection{Encoder Complexity}
% Each encoder layer processes the visible patch sequence $\mathbf{Z}^{(l-1)} \in \mathbb{R}^{2N_\text{v} \times d_\text{e}}$.
% We only analyze the dominant operations.

% \textit{Multi-head self-attention (MHSA):} 
% For each attention head $i \in \{1, \ldots, M_{\text{enc}}\}$, we compute the attention weights $\mathbf{A}_i$ as defined in Eq.~\eqref{eq:attention}. 
Each of query, key, and value projections requires $\mathcal{O}(\frac{2N_\text{v} d_\text{e}^2}{M_{\text{enc}}})$ operations for every attention head. Computing the attention matrix involves a matrix multiplication with complexity $\mathcal{O}\left(\frac{4N_\text{v}^2 d_\text{e}}{M_{\text{enc}}}\right)$. Applying attention to the result of value projection also adds $\mathcal{O}(\frac{4N_\text{v}^2d_\text{e}}{M_{\text{enc}}})$. Aggregating over all heads and including the output projection in Eq.~\eqref{eq:mhsa} with complexity $O(2N_\text{v} d_\text{e}^2)$ yields $\mathcal{C}_{\text{MHSA}} = \mathcal{O}(8N_\text{v}^2 d_\text{e} + 8N_\text{v} d_\text{e}^2)$. The two-layer MLP in Eq.~\eqref{eq:mlp} requires $\mathcal{O}(8N_\text{v} d_\text{e}^2)$. Combining $\mathcal{C}_{\text{MHSA}}$ and $\mathcal{C}_{\text{FFN}}$:

% \begin{equation}
% \mathcal{C}_{\text{MHSA}} = \mathcal{O}(8N_\text{v}^2 d_\text{e} + 8N_\text{v} d_\text{e}^2).
% \label{eq:c_mhsa}
% \end{equation}

% \textit{Feed-forward network (FFN):} 
% \begin{equation}
% \mathcal{C}_{\text{FFN}} = \mathcal{O}(2N_\text{v} d_\text{e} \cdot 2d_\text{e} + 2N_\text{v} \cdot 2d_\text{e} \cdot d_\text{e}) = \mathcal{O}(8N_\text{v} d_\text{e}^2).
% \label{eq:c_ffn}
% \end{equation}

% \textit{Total encoder complexity:}
% Combining $\mathcal{C}_{\text{MHSA}}$ from Eq.~\eqref{eq:c_mhsa}
% and $\mathcal{C}_{\text{FFN}}$ from Eq.~\eqref{eq:c_ffn} with negligible complexity $\mathcal{O}(2N_v d_e)$ of layer normalizations and residual additions:
\begin{equation}
\mathcal{C}_{\text{enc}} = L_{\text{enc}} \cdot (\mathcal{C}_{\text{MHSA}} + \mathcal{C}_{\text{FFN}}) = \mathcal{O}(L_{\text{enc}}(8N_\text{v}^2 d_\text{e} + 16N_\text{v} d_\text{e}^2)).
\label{eq:c_enc}
\end{equation}

The quadratic dependence on $N_\text{v}$ in the attention mechanism explains why high masking ratios significantly reduce computational cost. Reducing $N_\text{v}$ from $K$ to $(1-m_\text{r})K$ yields quadratic savings in the attention computation.

\subsection{Decoder Complexity}
%The decoder processes the complete sequence $\mathbf{Z}_{\text{full}} \in \mathbb{R}^{2K \times d_\text{e}}$ as defined in Eq.~\eqref{eq:z_full}, containing both encoded visible patches and learnable mask tokens. 
Following the same analysis as the encoder, but with a sequence length of $2K$ instead of $2N_\text{v}$, we have
\begin{equation}
\mathcal{C}_{\text{dec}} = \mathcal{O}(L_{\text{dec}}(8K^2 d_\text{e} + 16K d_\text{e}^2)).
\label{eq:c_dec}
\end{equation}

The decoder operates on the full sequence regardless of masking ratio, which motivates the asymmetric architecture where $L_{\text{dec}} \ll L_{\text{enc}}$. The output projection in Eq.~\eqref{eq:final_proj} %with $\mathbf{W}_{\text{out}} \in \mathbb{R}^{d_\text{e} \times d_\text{p}}$
incurs complexity $\mathcal{C}_{\text{proj}} = \mathcal{O}(2Kd_\text{e} d_\text{p})$.

\subsection{Contrastive Head Complexity}
% \textit{Mean pooling and MLP projection:} 
Computing $\mathbf{z}_{\text{mean}}$, as in Eq.~\eqref{eq:mean_pool}, requires $\mathcal{O}(2N_\text{v} d_\text{e})$ operations, whereas the contrastive projection in Eq.~\eqref{eq:mlp_contra} %with $\mathbf{W}^\text{C}_1 \in \mathbb{R}^{d_\text{e} \times 2d_\text{c}}$ and $\mathbf{W}^\text{C}_2 \in \mathbb{R}^{2d_\text{c} \times d_\text{c}}$ 
requires $\mathcal{O}(2d_\text{e} d_\text{c} + 2d_\text{c}^2)$, and normalizing the output embedding is $\mathcal{O}(d_\text{c})$. %\textit{Contrastive loss computation:}
% For batch size $\mathcal{B}$, 
Computing the contrastive loss % in Eq.~\eqref{eqn:contra_objective} 
over $2\mathcal{B}$ embeddings % $\mathbf{Z} \in \mathbb{R}^{2\mathcal{B} \times d_\text{c}}$
requires $\mathcal{O}(4\mathcal{B}^2 d_\text{c})$, leading to a total contrastive head complexity of:
% \begin{enumerate}
%     \item Computing pairwise similarities: $\mathbf{S} = \mathbf{Z}\mathbf{Z}^\top$ costs $\mathcal{O}(4\mathcal{B}^2 d_\text{c})$
%     \item Softmax over $2\mathcal{B}-1$ negative samples per anchor: $\mathcal{O}(4\mathcal{B}^2)$
%     \item Log and summation: $\mathcal{O}(2\mathcal{B})$
% \end{enumerate}

% Therefore, the complexity of contrastive loss calculation is $\mathcal{O}(4\mathcal{B}^2 d_\text{c})$, leading to a total contrastive head complexity of: 
\begin{equation}
\mathcal{C}_{\text{contra}} = \mathcal{O}(2N_\text{v} d_\text{e} + 2d_\text{e} d_\text{c} + 2d_\text{c}^2 + 4\mathcal{B}^2 d_\text{c}).
\label{eq:c_contra}
\end{equation}

The batch-dependent term $\mathcal{O}(4\mathcal{B}^2 d_\text{c})$ dominates for large batches, scaling quadratically with $\mathcal{B}$. This quadratic scaling presents a trade-off: while larger batches provide more negative samples that may improve contrastive learning quality, they incur substantial computational and memory costs.

\subsection{ContraWiMAE Complexity}
The overall pretraining complexity of ContraWiMAE with $N_\text{v} = (1-m_\text{r})K$ is
% \begin{align}
% \mathcal{C}_{\text{pretrain}} &= \mathcal{C}_{\text{prep}} + \mathcal{C}_{\text{enc}} + \mathcal{C}_{\text{dec}} + \mathcal{C}_{\text{proj}} + \mathcal{C}_{\text{contra}} \nonumber \\
% &= \mathcal{O}\Big(2(1-m_\text{r})Kd_\text{p} d_\text{e} \nonumber \\
% &\quad\quad + L_{\text{enc}}(8(1-m_\text{r})^2K^2 d_\text{e} + 16(1-m_\text{r})K d_\text{e}^2) \nonumber \\
% &\quad\quad + L_{\text{dec}}(8K^2 d_\text{e} + 16K d_\text{e}^2) + 2Kd_\text{e} d_\text{p} \nonumber \\
% &\quad\quad + 2(1-m_\text{r})K d_\text{e} + 2d_\text{e} d_\text{c} + 2d_\text{c}^2 + 4\mathcal{B}^2 d_\text{c}\Big).
% \label{eq:c_contrawimae}
% \end{align}

\begin{equation}    
\mathcal{C}_{\text{pretrain}} = \mathcal{C}_{\text{prep}} + \mathcal{C}_{\text{enc}} + \mathcal{C}_{\text{dec}} + \mathcal{C}_{\text{proj}} + \mathcal{C}_{\text{contra}}.
\end{equation}

During inference, only the encoder is used with $m_\text{r}=0$, so $N_\text{v} = K$ and therefore
\begin{align}
\mathcal{C}_{\text{inference}} &= \mathcal{C}_{\text{prep}} + \mathcal{C}_{\text{enc}} \nonumber \\
&= \mathcal{O}\Big(2Kd_\text{p} d_\text{e} + L_{\text{enc}}(8K^2 d_\text{e} + 16K d_\text{e}^2)\Big).
\end{align}

\subsection{Scalability implications:}
The complexity analysis reveals several key insights for scaling behavior. First, the $\mathcal{O}(K^2 d_\text{e})$ term in MHSA creates a quadratic attention bottleneck that becomes prohibitive for large channel matrices, full-attention becomes intractable without masking. Second, high masking ratios provide quadratic computational savings during pretraining without sacrificing inference capability. 
%Third, when only the encoder is deployed, the decoder depth can be increased to improve reconstruction quality without affecting the inference latency, as shown in Section \ref{sec:results}. 
Finally, while increasing $d_\text{e}$ improves representational capacity, it contributes $\mathcal{O}(d_\text{e}^2)$ to the computational cost of FFN layers. Therefore, a careful selection of $d_\text{e}$ is required to have a good balance between performance and efficiency. These complexity considerations guided our architectural choices and explain the performance-efficiency trade-offs observed experimentally.

Additionally, Table~\ref{tab:inference_performance} reports average run-time measurements of our method for both encoder-only (beam selection and LoS detection) and encoder-decoder (channel estimation) use cases. Consistent with our analysis, the results demonstrate improved efficiency at higher mask ratios, facilitating practical encoder-decoder deployment for reconstruction tasks. More importantly, the architectural design enables real-time operation across all considered tasks with low memory and computation footprints. 
Note that the encoder-only model has 12 layers with 16 attention heads and without masking, it takes more time and computation compared to the encoder-decoder model with masking and only 4 layers with 8 attention heads at the decoder.

\color{black}

\section{Analysis of Results}  \label{sec:results}

In this section, we examine in detail various model configurations and performance metrics across different downstream tasks and provide insight into optimal architectural design, pretraining strategies, \reviewer{robustness, }and fundamental relationships between model capacity, task complexity, and data efficiency. Our comprehensive evaluation reveals both the strengths of our approaches and the nuanced interplay among different architectural choices.

% \subsection{Optimal Architectural Design of WiMAE} 

\subsection{Comparison of WiMAE, ContraWiMAE, and Baselines}
\begin{figure}%[h]
    \centering
    \begin{subfigure}{0.99\columnwidth}
        \centering
        \includegraphics[width=0.99\columnwidth]{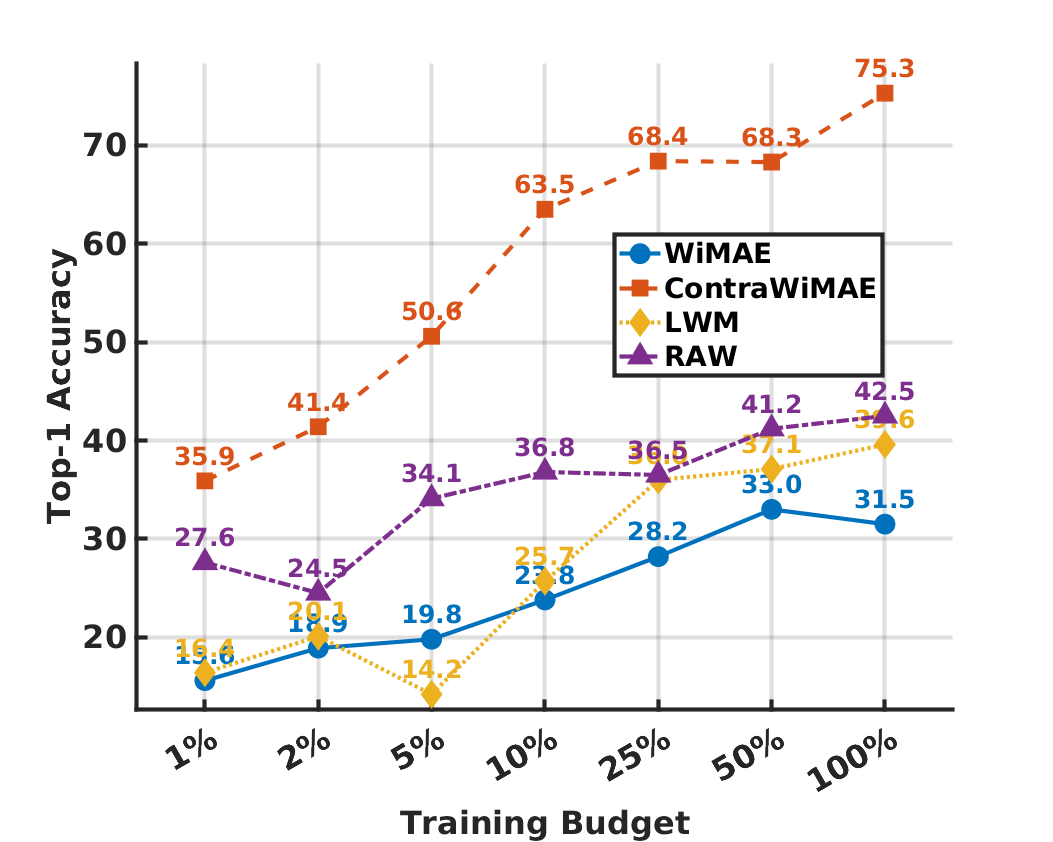}
    \caption{Beam selection with linear probing, codebook size 32.}
    \label{fig:linear_32}
    \end{subfigure}
    
    \vspace{0.7em}
    
    \begin{subfigure}{0.99\columnwidth}
        \centering
        \includegraphics[width=0.99\columnwidth]{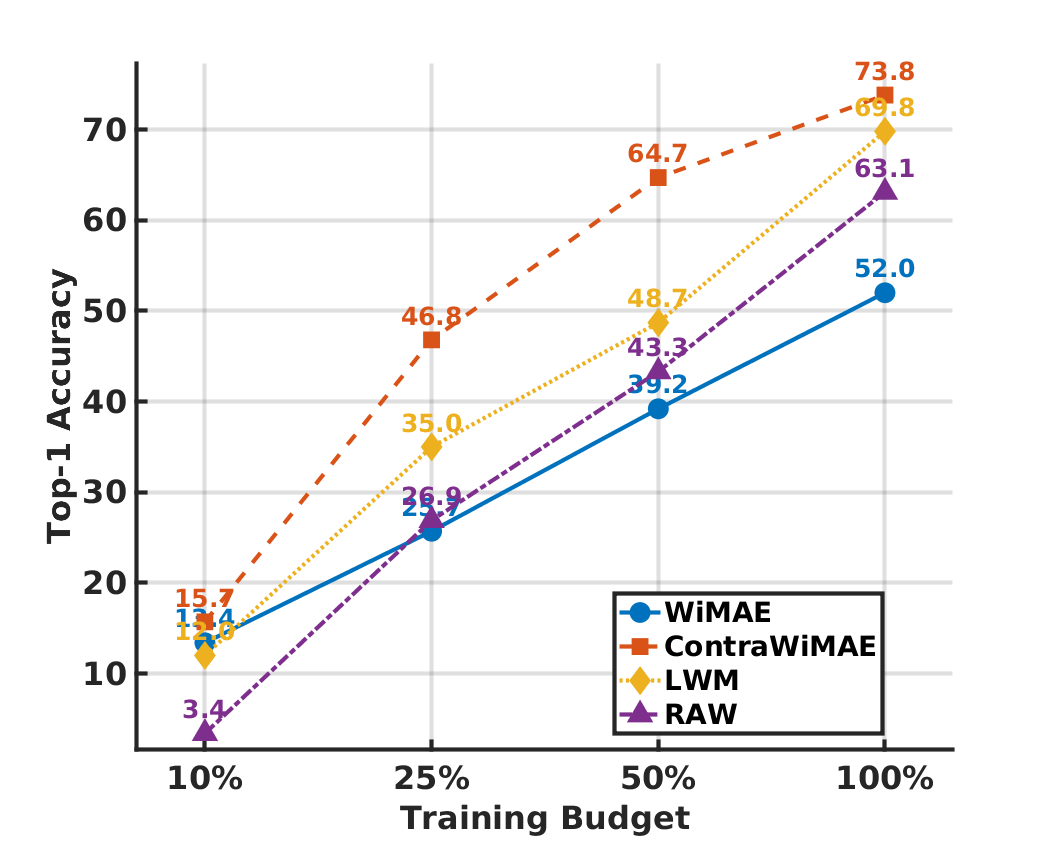}
        \caption{Beam selection with ResNet-Slim, codebook size 32.}
        \label{fig:resnet_slim}
    \end{subfigure}
    
    \vspace{0.7em}
    
    \begin{subfigure}{0.99\columnwidth}
        \centering
        \includegraphics[width=0.99\columnwidth]{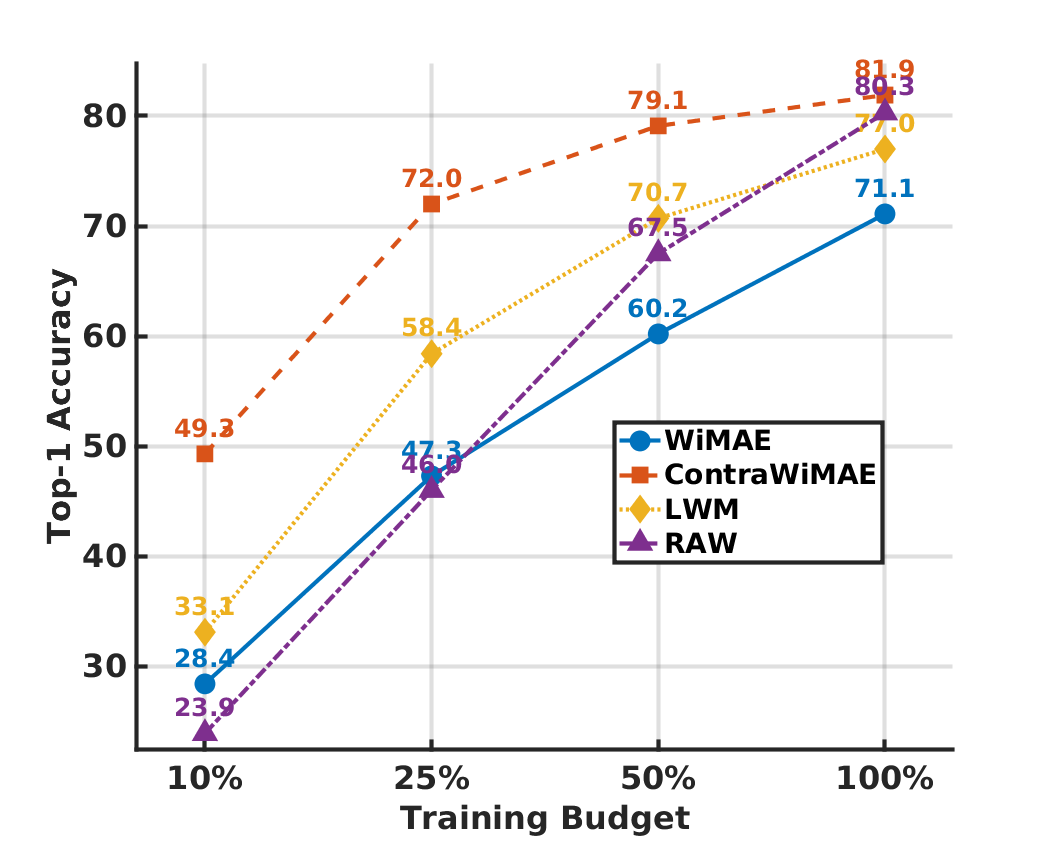}
        \caption{Beam selection with ResNet-Wide, codebook size 32.}
        \label{fig:resnet_wide}
    \end{subfigure}
    \caption{Top-1 beam selection accuracy for a codebook size of 32 achieved by ContraWiMAE vs. baselines followed by (a) linear probing, (b) ResNet-Slim, and (c) ResNet-Wide.}
    \label{fig:performance_analysis}
\end{figure}

\begin{figure*}[!t]
    \centering
    \includegraphics[width=\textwidth]{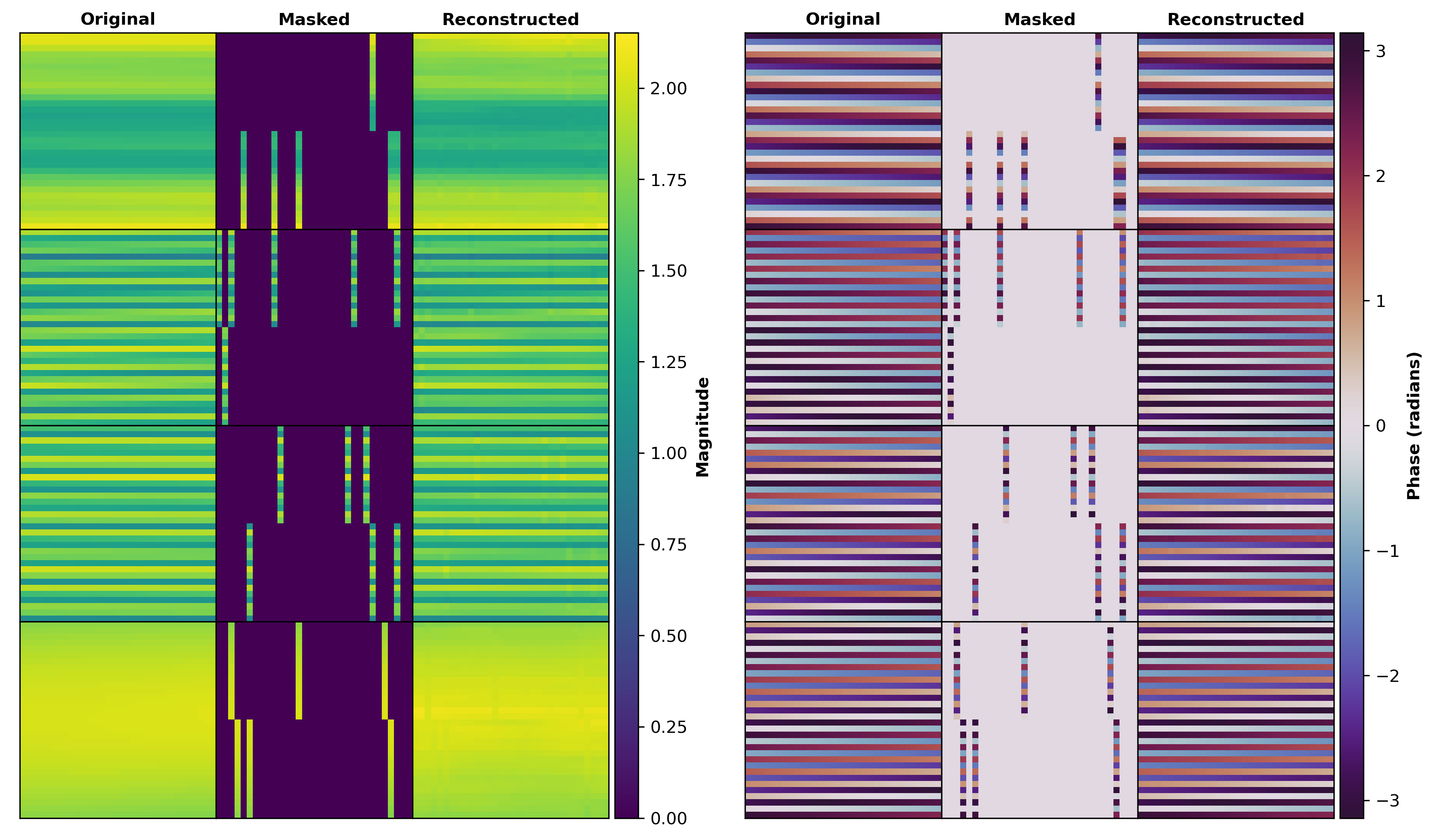}
    \caption{Magnitude (left) and phase (right) reconstruction for \reviewer{Contra}WiMAE, obtained with patch size $(N_{\text{p,s}}, N_{\text{p,f}})=\reviewer{(16,1)}$. Each row is a channel sample. Columns correspond to original, masked, and reconstructed channels.}
    \label{fig:rec}
\end{figure*}

In this section, we compare the performance of ContraWiMAE \reviewer{with those of WiMAE and} other baselines for beam selection \reviewer{(Tables~\ref{tab:linear_beam_selection}, \ref{tab:beam_selection_comparison}), } LoS detection \reviewer{(Table~\ref{tab:linear_los}), and channel estimation (Fig.~\ref{fig:ce_main})} downstream tasks. 
%The full evaluation is reported in Tables~\ref{tab:linear_beam_selection}, \ref{tab:beam_selection_comparison}, and \ref{tab:linear_los}. \reviewer{Channel estimation results are presented in Fig.~\ref{fig:ce_main}}, whereas 
\reviewer{In addition,} Fig.~\ref{fig:performance_analysis} highlights the accuracy vs. training budget tradeoff obtained on beam selection for a codebook size of 32 and different downstream models.
\reviewer{For the results in this section, we use a vertical patch shape with $(N_{\text{p,s}}, N_{\text{p,f}}) = (16, 1)$ and $m_\text{r}=0.9$ as shown in Fig. \ref{fig:rec}, since it accommodates the channel estimation task where we consider a pilot tone at a single subcarrier to estimate the channel.}

\subsubsection*{Linear probing on beam selection} 

The results in Fig.~\ref{fig:linear_32} demonstrate \reviewer{ContraWiMAE}'s data efficiency. For codebook size 32, with only 1\% training data, \reviewer{ContraWiMAE} achieves \reviewer{35.9}\% top-1 accuracy, surpassing \reviewer{all other baselines}. Notably, \reviewer{ContraWiMAE} with \reviewer{2}\% training data outperforms LWM trained on the entire dataset, demonstrating superior feature extraction capabilities. The addition of \reviewer{masked contrastive component} in ContraWiMAE \reviewer{over WiMAE} substantially enhances the linear separability of the encoded representations\reviewer{, highlighting that combining contrastive learning with masked autoencoding addresses fundamental limitations of WiMAE for wireless data.} With linear probing, as the training budget increases from 1\% to 100\%, ContraWiMAE demonstrates consistent performance scaling, with the advantage over \reviewer{other baselines} growing progressively larger. %From Table~\ref{tab:linear_beam_selection}, we observe that this scaling effect becomes more pronounced with complex tasks (i.e., larger beam codebook sizes), where the contrastive objective promotes a more homogeneous utilization of the latent space, enabling clearer decision boundaries even as class complexity increases. 
This pattern suggests that contrastive learning enables the model to \reviewer{gain substantial} discriminative power%even as classification complexity increases, 
, creating more linearly separable embeddings that preserve class distinctions without requiring complex downstream architectures.

\subsubsection*{ResNet models on beam selection}  

Figs.~\ref{fig:resnet_slim} and~\ref{fig:resnet_wide} examine the top-1 beam selection accuracy with a codebook size of 32 when the downstream model is a ResNet-Slim and a ResNet-Wide, respectively. While ContraWiMAE outperforms LWM by \reviewer{35.7}\% with linear probing, this gap narrows to just \reviewer{4.9}\% with ResNet-Wide, demonstrating how complex downstream architectures can offset weaker foundation representations. \reviewer{The diminishing gains with complex models like ResNet-Wide demonstrate the effectiveness of our pretraining. Complex downstream models can compensate for poor initial representations when given sufficient task-specific data, while our pretrained representations are already well-suited for the tasks, resulting in consistently strong performance without requiring extensive downstream model complexity.} Table~\ref{tab:resnet_beam_selection} further shows that ContraWiMAE maintains \reviewer{its} edge across \reviewer{all} configurations. %, especially in data-constrained scenarios. 
Notably, for %a codebook size of 16 and
\reviewer{small codebooks at} 100\% training budget, upgrading from linear probing to ResNet-Wide \reviewer{leads to minimal performance gain for ContraWiMAE, while for larger codebooks the difference becomes more pronounced.} %by only 4.8\%, while it improves LWM by 36.9\%. 
This indicates that \reviewer{ContraWiMAE}'s representations are already well optimized \reviewer{for simpler tasks}. For complex tasks, increased downstream capacity benefits all models.% though WiMAE maintains higher absolute performance.

\subsubsection*{Linear probing on LoS detection}  
Table~\ref{tab:linear_los} examines the performance of different models for LoS detection. %Both WiMAE and LWM 
\reviewer{All models} exceed 92\% accuracy even with just 1\% training data, suggesting that LoS characteristics are more easily extractable compared to beam indices. WiMAE maintains a slight advantage %with 100\% training data (96.0\% accuracy vs. 94.8\% for LWM)
\reviewer{over other models}. %WiMAE shows consistent improvement as the training data increases, while LWM exhibits small fluctuations, suggesting that WiMAE provides more stable feature spaces. 
ContraWiMAE shows competitive performance with WiMAE, while consistently outperforming LWM, suggesting that introducing the contrastive objective does not hurt the encoder's ability to produce meaningful compact summary representations. 

\begin{figure*}[t]
\centering
\begin{subfigure}[b]{0.3\textwidth}
    \centering
    \includegraphics[width=\textwidth]{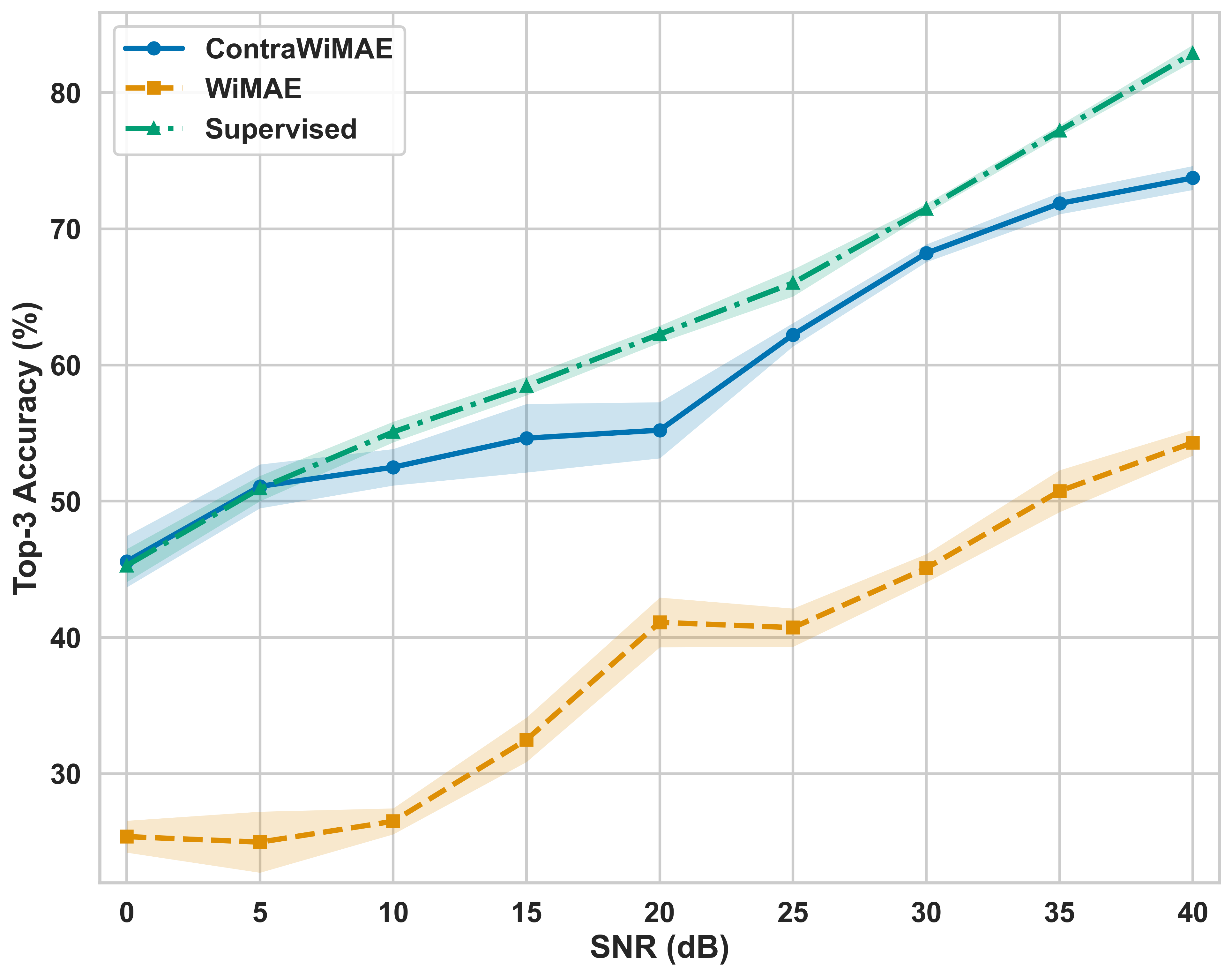}
    \caption{In-distribution, Beam size 16}
    \label{fig:indist_beam_16}
\end{subfigure}
\hfill
\begin{subfigure}[b]{0.3\textwidth}
    \centering
    \includegraphics[width=\textwidth]{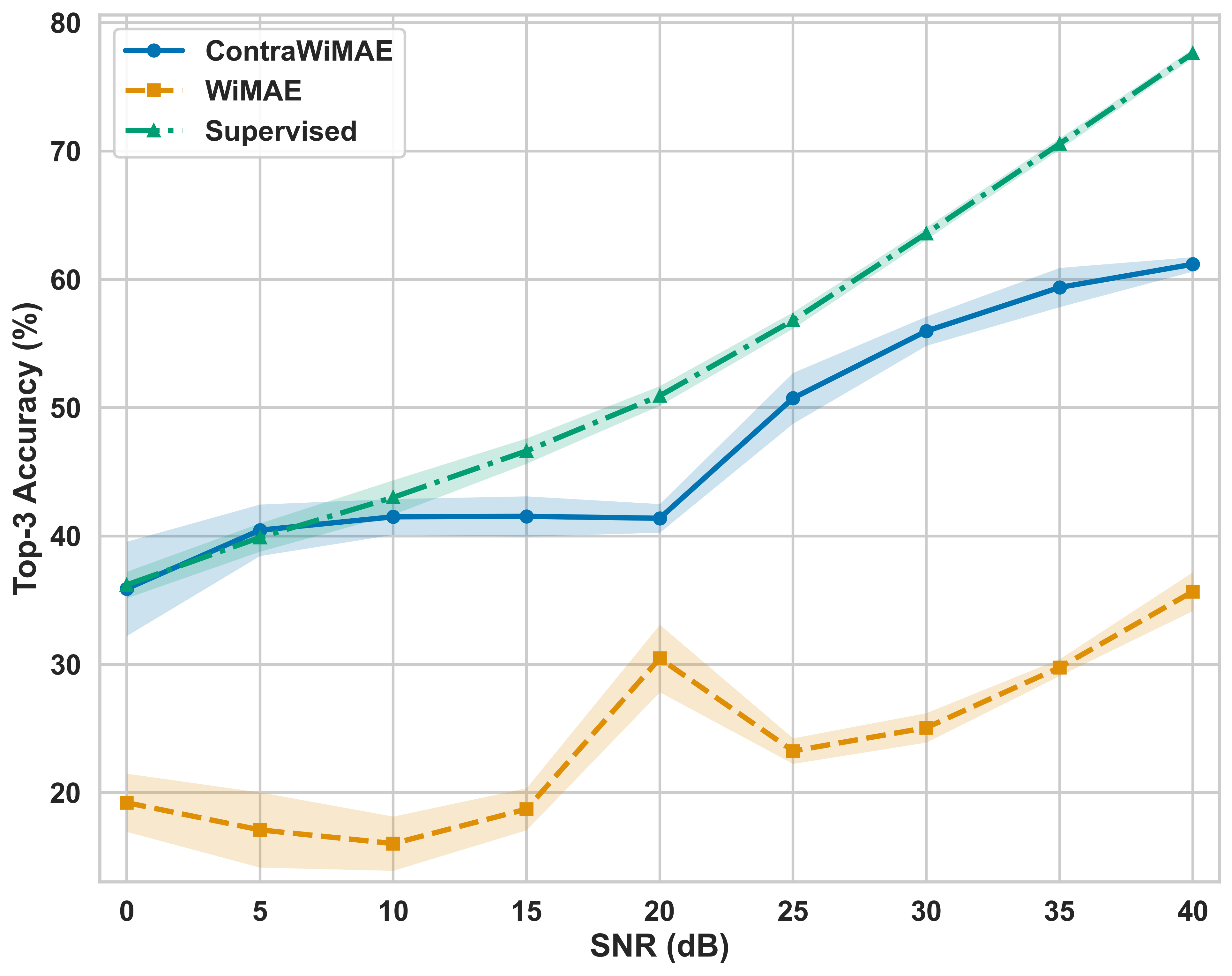}
    \caption{Same-frequency, Beam size 16}
    \label{fig:same_beam_16}
\end{subfigure}
\hfill
\begin{subfigure}[b]{0.3\textwidth}
    \centering
    \includegraphics[width=\textwidth]{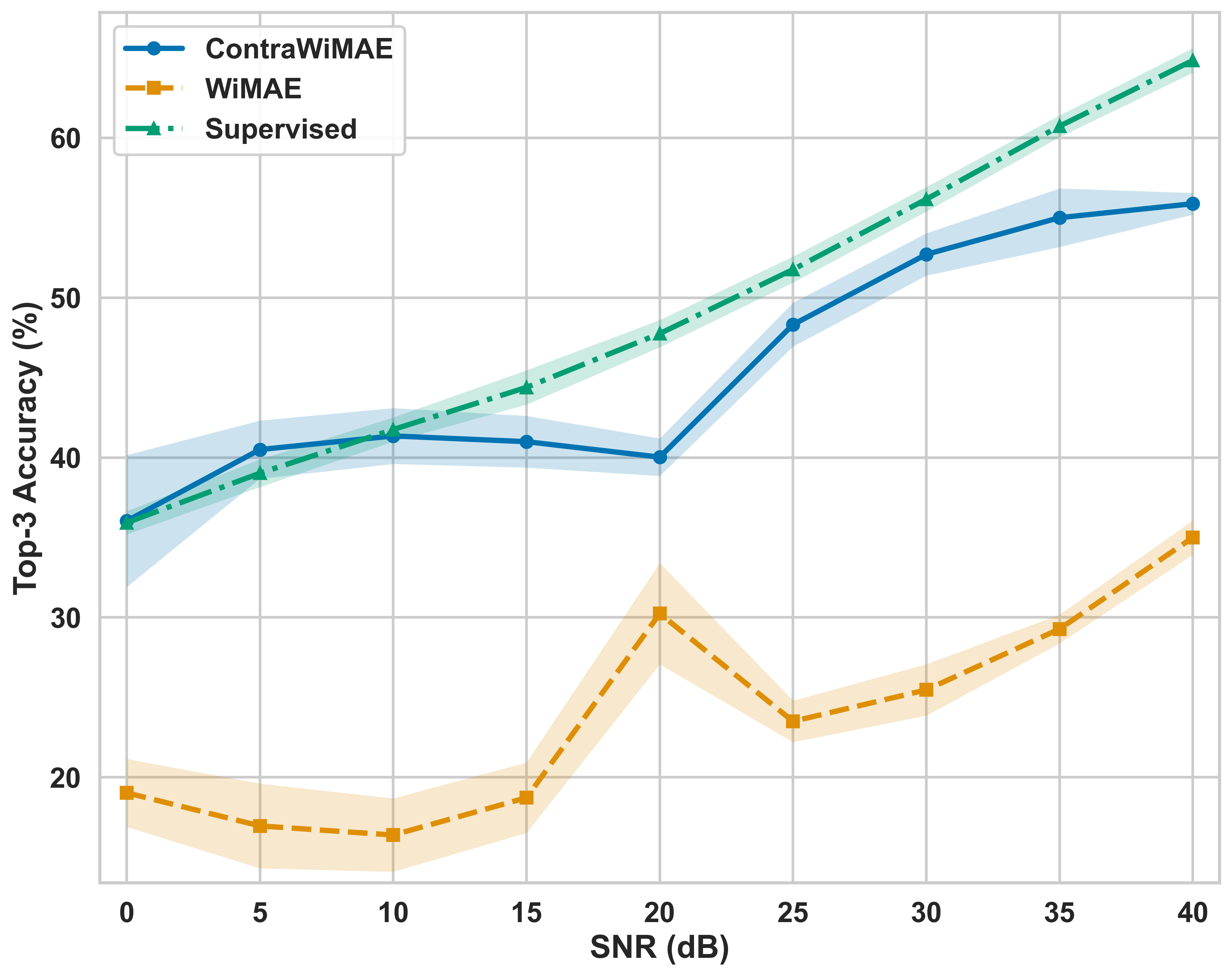}
    \caption{Cross-frequency, Beam size 16}
    \label{fig:cross_beam_16}
\end{subfigure}

\vspace{0.3cm}

\begin{subfigure}[b]{0.3\textwidth}
    \centering
    \includegraphics[width=\textwidth]{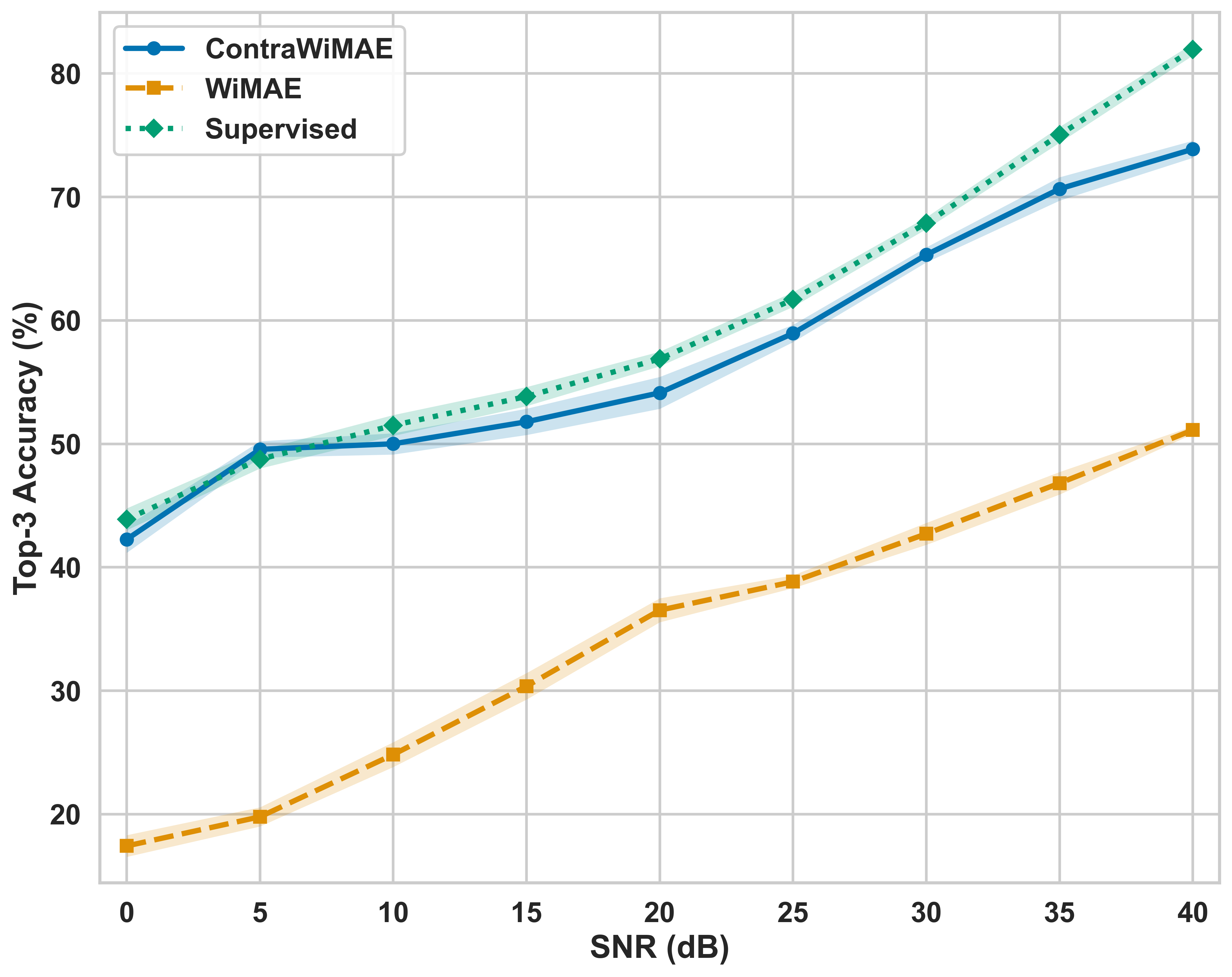}
    \caption{In-distribution, Beam size 32}
    \label{fig:indist_beam_32}
\end{subfigure}
\hfill
\begin{subfigure}[b]{0.3\textwidth}
    \centering
    \includegraphics[width=\textwidth]{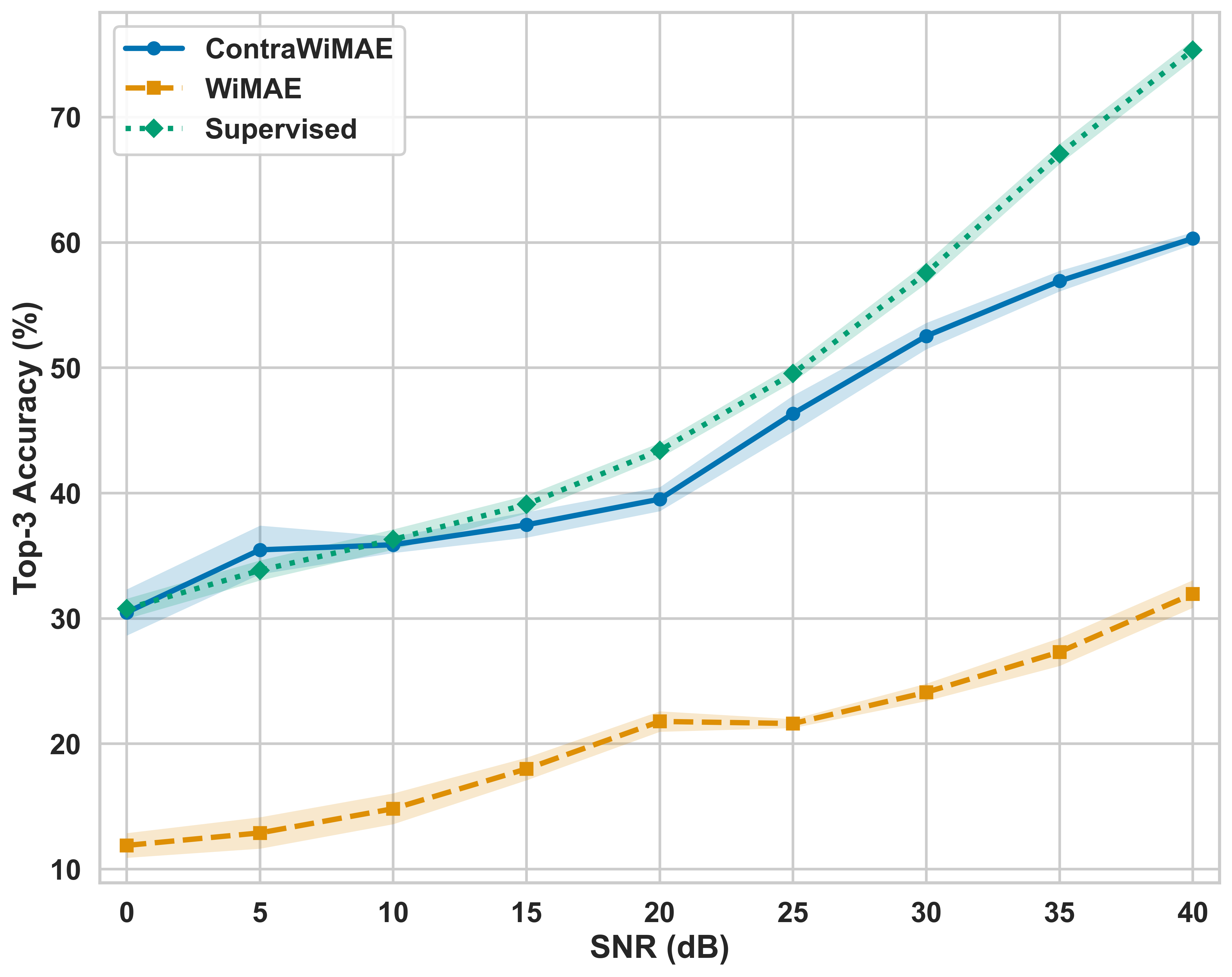}
    \caption{Same-frequency, Beam size 32}
    \label{fig:same_beam_32}
\end{subfigure}
\hfill
\begin{subfigure}[b]{0.3\textwidth}
    \centering
    \includegraphics[width=\textwidth]{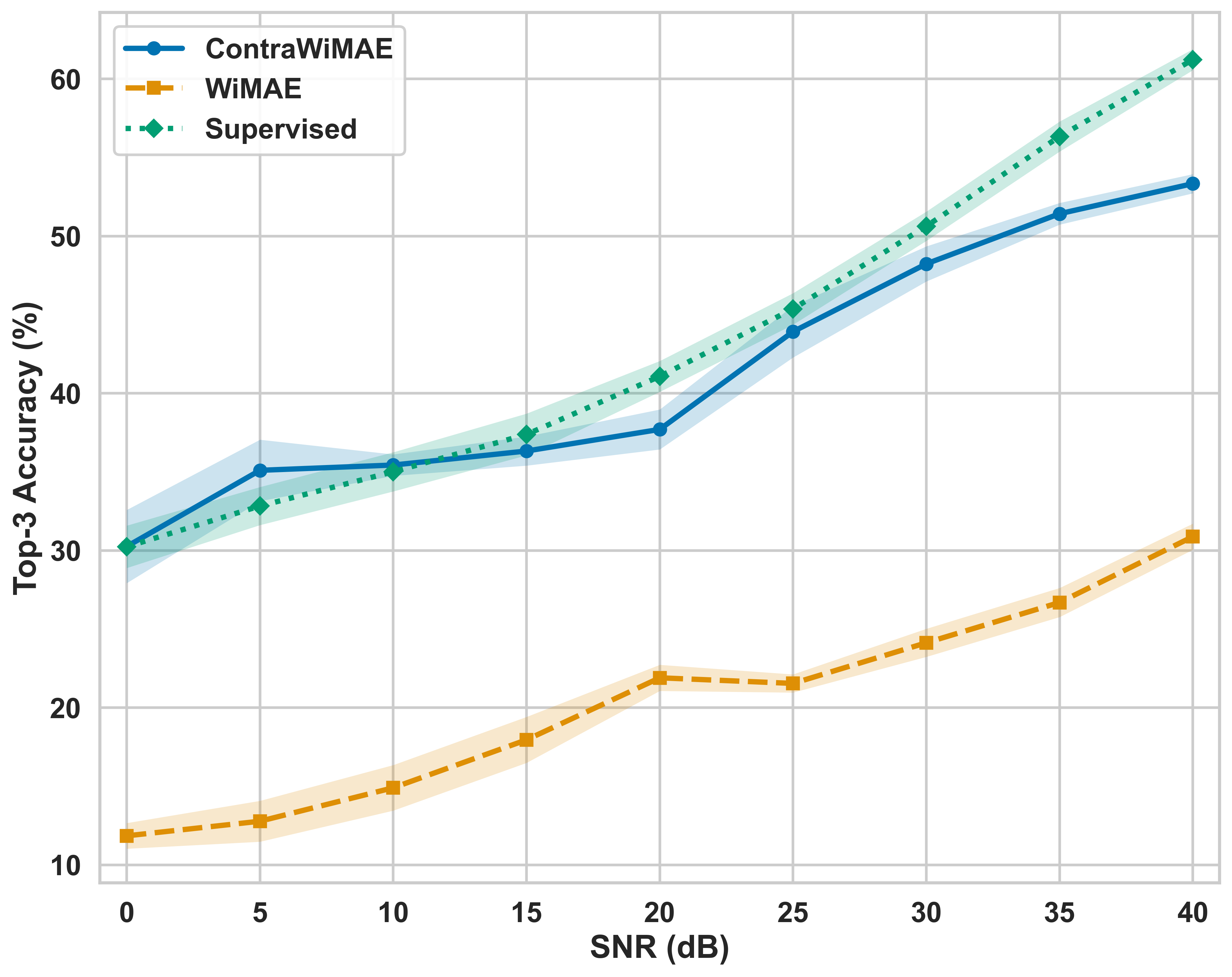}
    \caption{Cross-frequency, Beam size 32}
    \label{fig:cross_beam_32}
\end{subfigure}

\caption{Beam selection performance under varying SNR for in-distribution, same-frequency OOD, and cross-frequency OOD settings with beam sizes 16 (top row) and 32 (bottom row)}
\label{fig:beam_comparison}
\end{figure*}
\color{black}
\subsubsection*{Channel Estimation}
Although ContraWiMAE and WiMAE are pretrained under different scenarios with randomly applied masks, as shown in Fig.~\ref{fig:ce}, their performance remains comparable to the supervised model trained on the pretraining data with a single reconstruction mask that coincides with pilot positions. The dual objective of ContraWiMAE introduces only a slight reduction in reconstruction accuracy and its performance closely follows that of WiMAE. As expected, the supervised model trained with the same noise levels as those used during testing achieves the lowest NMSE, benefiting from its explicit exposure to noise characteristics during training.

Fig.~\ref{fig:ce_ft} further illustrates ContraWiMAE's scaling behavior with respect to the amount of finetuning data. The performance gains become less pronounced as the finetuning budget increases. Notably, when ContraWiMAE is finetuned on only 2.5\% of the task dataset corresponding to around 7000 channels, it surpasses all the supervised baselines across all SNR levels, highlighting its superior sample efficiency.

\subsubsection*{Reconstruction performance} 
Fig.~\ref{fig:rec} shows the reconstruction of both magnitude and phase information from heavily masked inputs using settings described in Section \ref{sec:meth:pretraining_of_wimae}. As shown in the figure, ContraWiMAE demonstrates impressive capability in capturing structural patterns rather than simply interpolating between known values. The reconstructions preserve characteristic fading patterns, spatial correlations across antenna elements, and coherent phase relationships between antenna elements and subcarriers. The transitions between visible and reconstructed regions are smooth, without boundary artifacts, demonstrating spatial coherence and the model's ability to leverage global context. This quality of reconstruction with \reviewer{90}\% masking suggests that ContraWiMAE learns meaningful representations of the underlying channel physics rather than memorizing patterns, making it well-suited for fine-tuning on task-specific scenarios.

\subsection{Robustness Analysis}
We study the ContraWiMAE's robustness by examining the beam selection performance of a simple kNN classifier trained on encoder representations obtained under various distribution shift conditions as shown in Fig.~\ref{fig:beam_comparison}. Under in-distribution evaluation, where the train and test points of kNN are from the same downstream task scenarios and the beam labels are also obtained at 3.5\,GHz, we evaluate the noise robustness of ContraWiMAE. Figs.~\ref{fig:indist_beam_16} and \ref{fig:indist_beam_32} show that ContraWiMAE progressively narrows the gap with supervised approaches. Remarkably, in the most challenging 0-10\,dB SNR regime, our method achieves parity with supervised training. We next assess the generalization to out-of-distribution (OOD) scenarios through cross-scenario evaluation (Figs.~\ref{fig:same_beam_16} and \ref{fig:same_beam_32}), where test environments remain completely separate from those used in kNN training. Here, ContraWiMAE retains its advantage over supervised baseline at lower SNR levels, with all methods exhibiting comparable degradation.

The most demanding evaluation appears in Figs~\ref{fig:cross_beam_16} and \ref{fig:cross_beam_32}, where we simultaneously introduce both unseen test scenarios and unseen carrier frequency of 28\,GHz, creating both an environmental and spectral shift. Despite operating without labeled supervision, ContraWiMAE's inherent noise robustness enables it to surpass the supervised baseline when these compounded shifts coincide with adverse noise conditions.

\begin{figure*}
    \centering
    \includegraphics[width=0.75\linewidth]{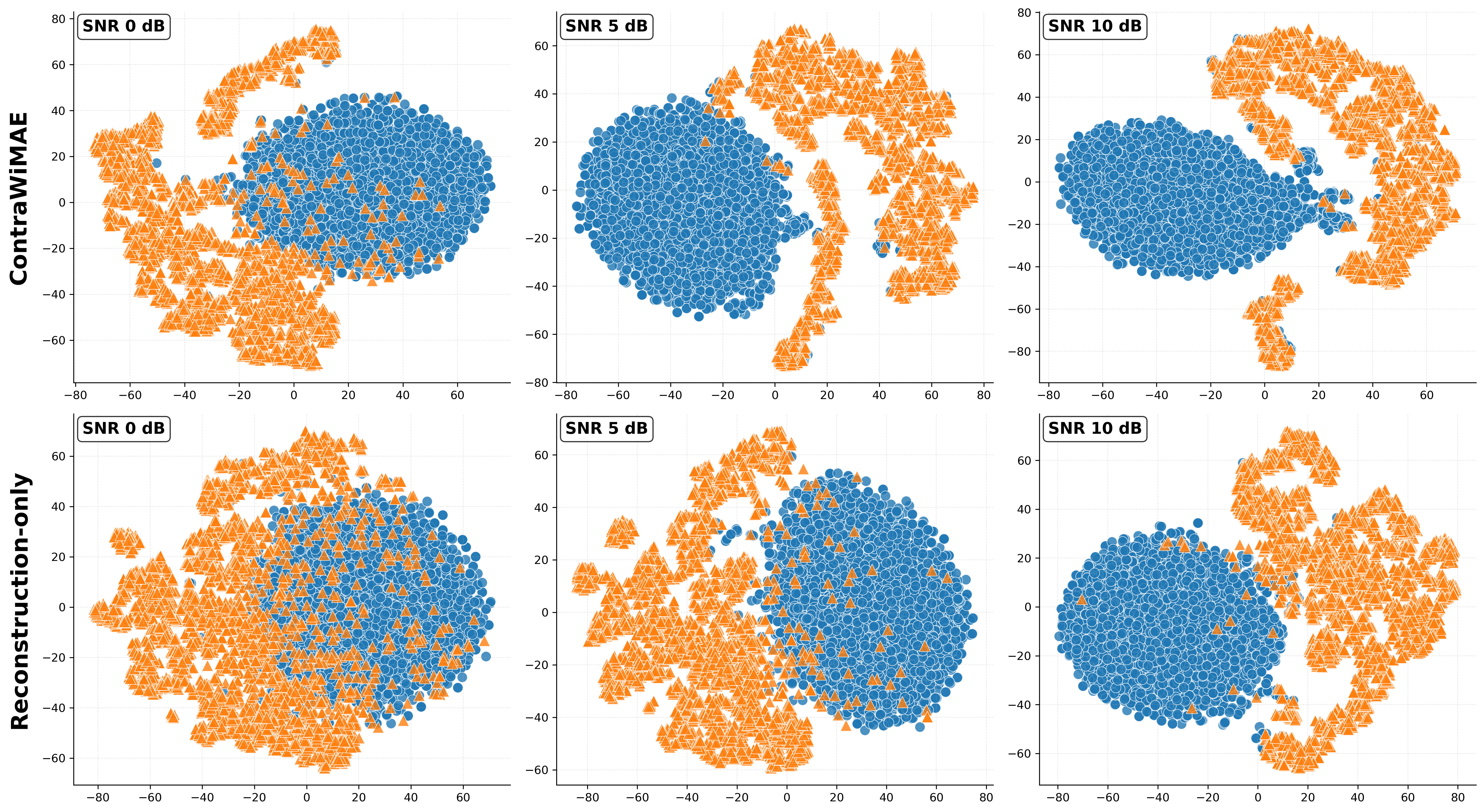}
    \caption{t-SNE representation with LoS (blue) and non-LoS (orange) labels under varying SNR}
    \label{fig:tsne}
\end{figure*}

Additionally, we present t-SNE visualizations in Fig.~\ref{fig:tsne} to illustrate the robustness of ContraWiMAE compared to WiMAE under varying signal-to-noise ratio conditions (0, 5, and 10\,dB). The representations are differentiated according to the presence of a LoS propagation path. While both approaches demonstrate satisfactory cluster separation at 10\,dB SNR, ContraWiMAE exhibits substantially more distinct class boundaries under degraded signal conditions (0 and 5\,dB SNR). Conversely, WiMAE demonstrates considerable misclassification, frequently confounding non-LoS channels with their LoS counterparts under these adverse conditions. This observation underscores the superior noise resilience conferred by the masked contrastive objective.

\color{black}

\subsection{Sensitivity to Architectural Design Choices}
\begin{figure}%[htbp]
    \centering
    \begin{subfigure}{0.99\columnwidth}
        \centering
        \includegraphics[width=0.95\columnwidth]{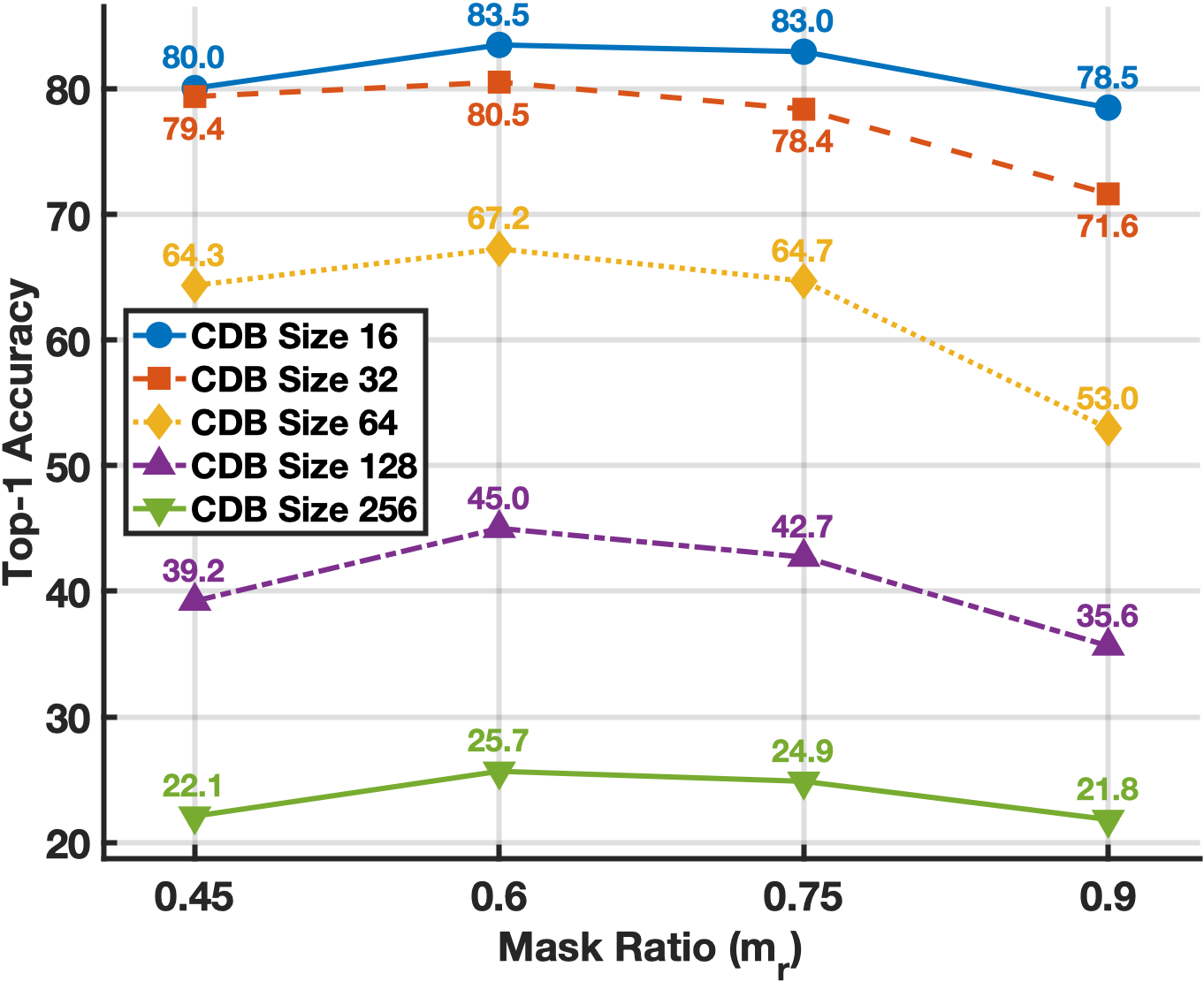}
        \caption{Top-1 accuracy vs. mask ratio.}
        \label{fig:top1_mr}
    \end{subfigure}
    
    \vspace{0.7em}
    
    \begin{subfigure}{0.99\columnwidth}
        \centering
        \includegraphics[width=0.95\columnwidth]{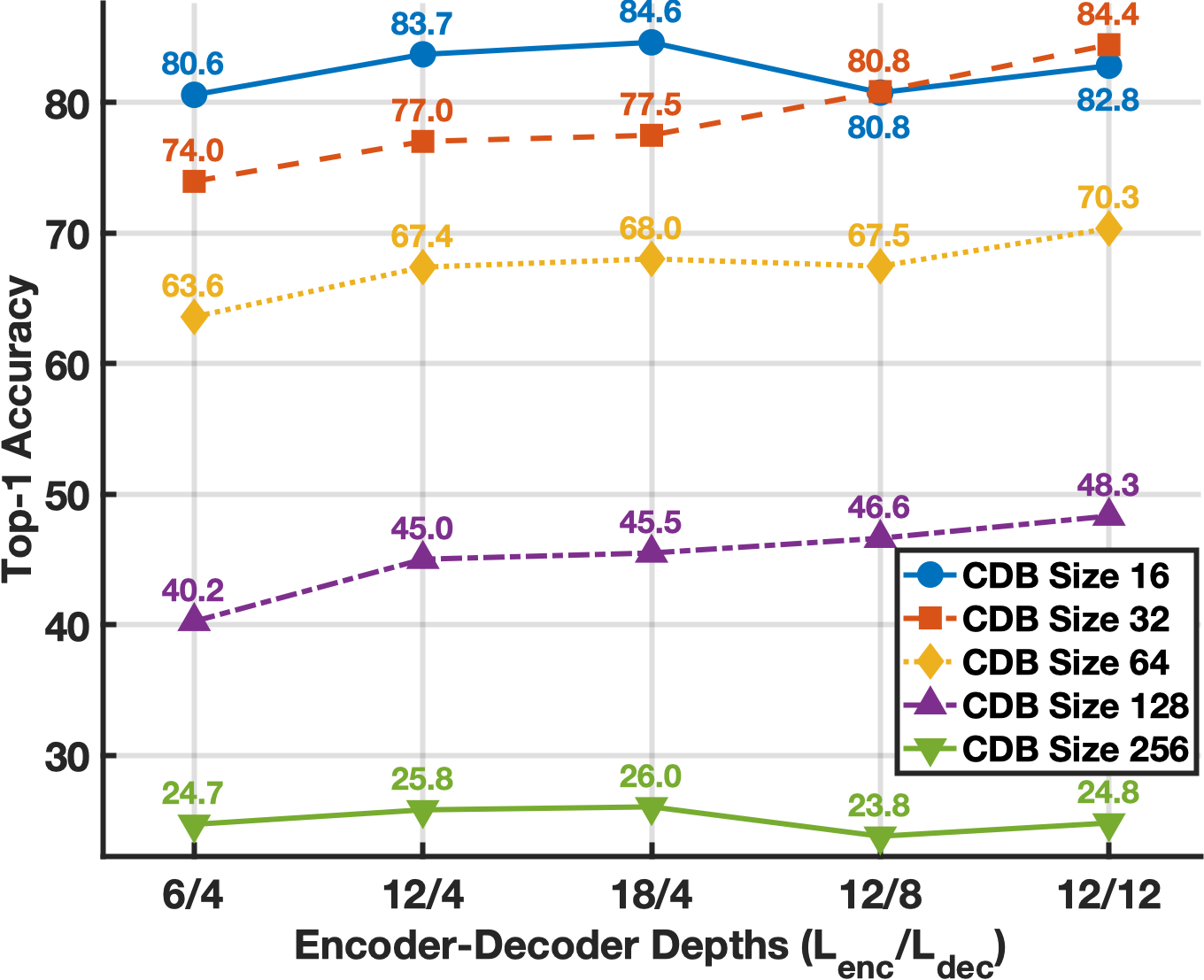}
        \caption{Top-1 accuracy vs. model depth.}
        \label{fig:top1_depth}
    \end{subfigure}
    
    \vspace{0.7em}
    
    \begin{subfigure}{0.99\columnwidth}
        \centering
        \includegraphics[width=0.95\columnwidth]{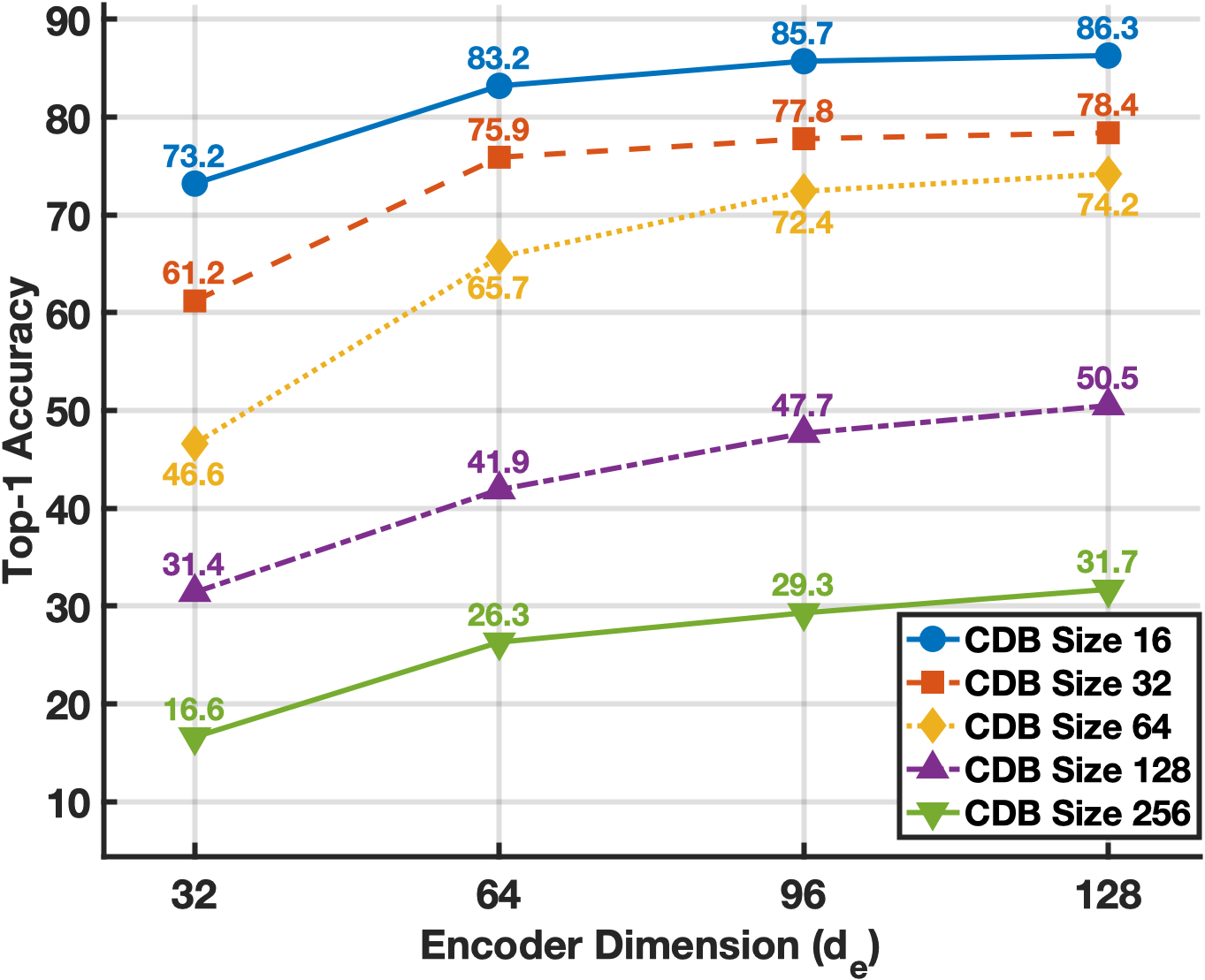}
        \caption{Top-1 accuracy vs. dimension.}
        \label{fig:top1_dim}
    \end{subfigure}
    \caption{Beam selection performance for WiMAE with linear probing across different model parameters and codebook (CDB) sizes.}
    \label{fig:hyperparam_search}
\end{figure}
 \reviewer{To analyze the sensitivity of the model behavior to architecture design choices, we explore different masking ratios $m_{\text{r}} \in \{0.45, 0.60, 0.75, 0.90\}$, network depths $L_{\text{enc}} \in \{6, 12, 18\}$ and $L_{\text{dec}} \in \{4,8,12\}$, encoding dimensions $d_{\text{e}} \in \{32, 64, 96, 128\}$, and patch configurations $(N_{\text{p,s}}, N_{\text{p,f}}) \in \{(16, 1), (1,16), (4,4)\}$, with fixed attention heads $M_{\text{enc}}=16$ and $M_{\text{dec}}=8$. For example, in Fig.~\ref{fig:hyperparam_search} that  summarizes the sensitivity results, we use a different patch size $(N_{\text{p,s}}, N_{\text{p,f}})$ = (1, 16).
We begin by analyzing the impact of the masking ratio, the depth of the model, and the representation dimension on the performance of the WiMAE, providing insights into the sensitivity to design choices for reconstruction-only foundation. To this end, we assess the beam selection task performance along with the reconstruction loss from Eq.~\eqref{eq:recon_loss}, which we minimize during pretraining.} 
\reviewer{The architectural choices are critical as they determine the trade-off between computational efficiency and representation quality, which are essential considerations for practical deployment in resource-constrained wireless systems. While our sensitivity analysis focuses on the specific tasks and datasets considered in this work, it provides practical insights into the mechanistic characteristics of our method and guidance for architectural design decisions.} 

% We also provide insights into the masked channel reconstruction, which is at the base of the \reviewer{reconstructive} pretraining stage. 

\subsubsection*{Masking ratio}

\reviewer{Fig.~\ref{fig:top1_mr} reveals the interplay between mask ratio and feature quality, reflecting the need to balance task difficulty; a small ratio makes reconstruction trivial while a large ratio may render the learning intractable. This suggests an optimal information density for the beam selection task where sufficient masking forces the model to learn meaningful representations without compromising its ability to capture essential discriminative channel characteristics. Moreover, we observe that the same phenomenon applies to reconstruction tasks where the performance is highly dependent on finding a suitable mask ratio. Note that task difficulty is not only determined by the mask ratio but also by patch shape, which plays an important role. We found $m_\text{r}=0.6$ to be a fair trade-off point for a patch shape of $(N_\text{p,s}, N_\text{p,f})=(1,16)$. 
}

\subsubsection*{Architecture depth}  
Fig.~\ref{fig:top1_depth} demonstrates the complex relationship between model depth and performance \reviewer{for the task and dataset considered}. As the encoder depth increases from 6 to 12 layers, we observe consistent performance improvements across all codebook sizes. However, the marginal benefit diminishes beyond 12 layers, with 18-layer models showing minimal improvement. This suggests that moderate encoder depth effectively captures channel characteristics without excessive computation or potential overfitting. More interestingly, for a fixed 12-layer encoder, increasing the decoder depth from 4 to 8 and then to 12 layers provides substantial performance improvements. When the decoder is discarded after pretraining, its added complexity does not affect inference latency. This asymmetric architecture---moderate encoder with deep decoder---reveals that a sophisticated decoding process \reviewer{may} provide a stronger learning signal that helps the encoder capture more nuanced channel features during training.

\subsubsection*{Representation dimension} 
Fig.~\ref{fig:top1_dim} reveals a strong positive correlation between embedding dimension and performance, with patterns that vary by task complexity. Top-1 accuracy improves as the dimension increases from 32 to 128, with diminishing returns becoming apparent beyond 96. The most significant improvement occurs between dimensions 32 and 64, suggesting that this range is critical for adequately representing channel characteristics for the data under consideration. More challenging tasks show less saturation at higher dimensions, indicating that complex discrimination tasks \reviewer{may} benefit more from increased representational capacity.

\color{black}

\section{Conclusion} \label{sec:conclusion}

In this paper, we introduced \reviewer{ContraWiMAE, a transformer-based contrastive masked autoencoder,} specifically designed for wireless channel representation learning. By employing an asymmetric encoder-decoder architecture \reviewer{with} high masking ratios and moderate encoder depth, \reviewer{ContraWiMAE} strikes a balance between representational quality and computational efficiency. \reviewer{The key innovation lies in our novel masked contrastive objective exploiting the inherent characteristics of wireless propagation by applying distinct random masks to noisy observations of identical channel realizations. This approach enables simultaneous extraction of both coherent structural patterns and discriminative characteristics.}

%To further enhance the quality of learned representations, we proposed ContraWiMAE, which extended WiMAE by incorporating a contrastive learning objective in a unified multi-task framework. Warm-starting from pretrained WiMAE weights and using additive noise to generate positive pairs, ContraWiMAE captures both structural and discriminative features of wireless channels. This hybrid approach further improves the linear separability of the learned embeddings, leading to enhanced downstream performance, especially with simple classifiers. Importantly, ContraWiMAE preserves the strong reconstruction performance of its base model, demonstrating that the two learning objectives can effectively complement each other.

Our experiments \reviewer{demonstrated} that \reviewer{ContraWiMAE} consistently outperforms existing baselines across multiple downstream tasks, achieving particularly strong results under constrained data conditions\reviewer{, a capability that is especially} critical for practical wireless scenarios where labeled data are expensive or difficult to obtain. Through extensive evaluation on realistic, unseen wireless scenarios, ContraWiMAE exhibited strong \reviewer{cross-frequency and cross-environment} transferability\reviewer{, as well as noise} robustness. We showed that, unlike other models that require complex downstream architectures to achieve good performance, ContraWiMAE produces representations that remain effective even with lightweight classifiers. \reviewer{We validated our findings across tasks requiring both discriminative understanding (LoS detection and beam selection) and reconstruction capability (channel estimation). % We also demonstrated performance that matches and surpasses supervised baselines under challenging conditions.
Under challenging low-SNR conditions and distribution shifts, ContraWiMAE matches or surpasses supervised baselines, validating robustness of our wireless-inspired augmentation strategy.}

\reviewer{Beyond technical contributions, this work contributes to a paradigm shift toward unified wireless AI systems where a single foundation model serves multiple applications with minimal adaptation overhead. Future work could extend this framework to MIMO configurations, investigate its use for other downstream tasks, and develop efficient compression techniques for edge deployment. Another extension is to include mobility and Doppler effects by analyzing time-domain correlations to capture temporal channel dynamics. While our current evaluation uses ray-tracing data with idealized antenna patterns, future extensions could incorporate hardware impairments and validation on over-the-air measurements to further demonstrate the framework's robustness in operational deployments. We envision ContraWiMAE serving as a strong baseline, providing a principled approach to self-supervised wireless channel representation learning that respects the unique characteristics of wireless propagation while achieving practical performance across diverse applications. 
We release the ContraWiMAE model weights and training pipeline to support reproducibility and future advances in self-supervised wireless channel representation learning.}
%WiMAE achieved optimal performance at a masking ratio of 0.6, with deeper decoders offering better representations without affecting inference latency. Larger embedding dimensions were found to be especially beneficial for complex tasks, and the data efficiency of both models enables high performance using as little as 1\% of the training data employed by other baselines.

\bibliographystyle{IEEEtran}
\bibliography{references}
\end{document}